\documentclass{article}
\usepackage{log_2026}                        

\ifdefined\nolinenumbers\nolinenumbers\fi

\usepackage{booktabs}
\usepackage{multirow}
\usepackage{amsfonts}
\usepackage{amssymb}
\usepackage{graphicx}
\usepackage{xcolor}
\usepackage{colortbl}
\definecolor{relgtblue}{RGB}{0,172,219}
\usepackage[normalem]{ulem}
\usepackage{mathtools}
\usepackage{float}
\usepackage{algorithm}
\usepackage{algpseudocode}
\usepackage{enumitem}
\usepackage[numbers,compress,sort]{natbib}
\usepackage{pdflscape}
\usepackage{caption}

\newcommand{\seed}{s}
\newcommand{\tseed}{t_{\mathrm{seed}}}
\newcommand{\Nval}{\mathcal{N}}

\newcommand{\hbfs}{h_{\mathrm{BFS}}}
\newcommand{\hwalk}{h_{\mathrm{walk}}}
\newcommand{\Fsketch}{F_{\mathrm{sketch}}}

\title[Published as a conference paper at Learning on Graphs (LoG) 2026]{\textsc{Quartet}: QUad-branch cross-Attention and Random-walk Traces for Enhancing Transformers on Relational Graphs}

\author[Myint et al.]{%
  \normalsize
  \makebox[\textwidth][c]{%
    \textbf{Kyaw Hpone Myint\textsuperscript{*}}\hfil
    \textbf{Nan Jiang\textsuperscript{\dag}}\hfil
    \textbf{Xiang Li\textsuperscript{\dag}}%
  }\\[0.3em]
  \normalsize
  \makebox[\textwidth][c]{%
    \textbf{Zhe Wu\textsuperscript{\dag}}\hfil
    \textbf{Alexandre G.R.\ Day}\hfil
    \textbf{Pranab Mohanty}\hfil
    \textbf{Giri Iyengar}%
  }\\[0.6em]
  Capital One\\
  \email{\{kyaw.hponemyint, nan.jiang, xiang.li, zhe.wu}\\
  \email{alexandre.day, pranab.mohanty, giridharan.iyengar\}@capitalone.com}%
}

\begin{document}

\maketitle
\renewcommand{\thefootnote}{\fnsymbol{footnote}}
\footnotetext[1]{First and corresponding author.}
\footnotetext[2]{Equal contribution second authors.}
\renewcommand{\thefootnote}{\arabic{footnote}}

\begin{abstract}
Relational Deep Learning (RDL) models multi-table databases as heterogeneous temporal graphs, and graph transformers currently achieve state-of-the-art performance on benchmarks like \textsc{RelBench}.
However, the current leading model, RelGT, suffers from two key limitations: its random local sampler yields loosely connected subgraphs that hinder message passing, and its global attention module relies on a single, seed-feature-based memory that ignores broader macro-level dynamics.
To overcome these limitations, we introduce \textsc{Quartet}, an expressive graph transformer architecture that applies full self-attention on local subgraphs while enriching global context through cross-attention branches.
Specifically, \textsc{Quartet} employs a Causal Random Walk (CRW) sampler based on recency-truncated Personalized PageRank (PPR) to extract compact, hub-robust, and densely connected local subgraphs without temporal leakage.
Concurrently, a quad-branch cross-attention module integrates global context from four complementary perspectives: seed feature, seed topology, temporal dynamics, and collaborative dynamics.
Across the \textsc{RelBench} v1 classification tasks, \textsc{Quartet} consistently matches or outperforms the current state-of-the-art graph transformer baselines (HGT and RelGT).
Ablation studies confirm that the CRW sampler significantly enriches local neighborhood quality, while the global branches provide essential, task-specific predictive gains.
\end{abstract}

\section{Introduction}

\emph{Relational databases} (RDBs)~\citep{codd1970relational} form the backbone of industrial machine learning by archiving extensive real-world datasets, such as financial logs and medical records. Through their interconnected, multi-table architectures, RDBs leverage primary-foreign key dependencies to explicitly trace how different entities engage and change over time.
Predictive tasks over these databases, such as churn, sales forecasting, and recommendation, are central to modern enterprise machine learning.
Yet extracting predictive signal from this data has historically required flattening its multi-table structure into a single feature table through labor-intensive, error-prone feature engineering~\citep{kanter2015deep} before handing it to tabular learners such as gradient-boosted decision trees~\citep{chen2016xgboost,ke2017lightgbm}.

\emph{Relational Deep Learning} (RDL) removes this bottleneck by treating an RDB as a heterogeneous temporal graph~\citep{trivedi2019dyrep,xu2020tgat,rossi2020tgn} (rows become nodes and primary-foreign-key links become edges) and learning end-to-end over the resulting \emph{relational entity graph} (REG)~\citep{fey2024position,dwivedi2025rdl,chen2025relgnn,yuan2024contextgnn,lachi2025lightrdl}.
The standard RDL model is a heterogeneous message-passing GNN with time-aware neighbor sampling and per-type tabular encoders~\citep{hamilton2017graphsage,kipf2017gcn,schlichtkrull2018rgcn,wang2019han}.
While such models capture local structure well, they inherit the documented limitations of message passing: bounded expressiveness~\citep{xu2019gin,morris2019weisfeiler,loukas2020depth} and an inability to model long-range dependencies without oversquashing information through narrow bottlenecks~\citep{alon2021bottleneck}.
These limits are acute on REGs, where two rows in the same table interact only through multi-hop paths that traverse shared parents: a shallow GNN cannot connect them, yet stacking layers to reach them induces over-smoothing.

\emph{Graph Transformers} (GTs) circumvent these topological constraints by replacing fixed-hop message passing with attention that reasons across structurally distant parts of the sampled subgraph~\citep{dwivedi2021generalization,ying2021graphormer,rampasek2022graphgps,kreuzer2021san,chen2022sat,chen2023nagphormer,shirzad2023exphormer,wu2023sgformer,kong2023goat,mialon2021graphit,mao2023hinormer,velickovic2018gat}.
GTs have thus emerged as a leading supervised RDL model, topping the standardized \textsc{RelBench} benchmark~\citep{robinson2024relbench,gu2026relbench}.
Notable architectures include the Heterogeneous Graph Transformer (HGT)~\citep{hu2020hgt}, which accommodates relational data using meta-relation-dependent attention and relative temporal encodings.
The Relational Graph Transformer (RelGT)~\citep{dwivedi2026relgt} augments a local-subgraph transformer with a global module that queries \emph{learnable centroids} maintained by an EMA K-Means clustering. 
Meanwhile, the Relational Graph Perceiver (RGP)~\citep{lachi2025rgp} compresses structural and temporal context through a Perceiver cross-attention bottleneck, making time an active modeling signal rather than just a filtering constraint.

Building upon these advances, our work addresses two specific limitations in the strongest GT baseline, RelGT.
First, its local sampling yields a loosely connected subgraph: RelGT populates each seed's neighborhood by drawing one- and two-hop nodes largely at random, so many sampled nodes end up with no edge to the seed (or to one another) inside the subgraph, leaving the local transformer to pass messages across a fragmented, sparsely linked set of nodes.
Second, its global context is narrow: RelGT's global module attends only over the seed's features, with no view of the seed's surrounding topology or of broader signals such as which entities are active at the same time (temporal context) or behave in similar ways (collaborative context), so relevant context that lives in structurally distant parts of the graph never reaches the model.
The first gap motivates our Causal Random Walk sampler, which concentrates the local subgraph on structurally relevant neighbors; the second motivates our quad-branch global module, which integrates feature, topological, temporal, and collaborative contexts to enrich the graph representation.

\paragraph{Contributions.}
To address these challenges, we introduce \textsc{Quartet}, a unified architecture that overcomes the limitations of existing relational deep learning models by seamlessly integrating enriched, localized structural dynamics with global, structurally distant graph contexts.
Concretely, our main contributions are:
\begin{enumerate}
  \item \textbf{A Causal Random Walk (CRW) Sampler for Enriching Local Subgraphs:} We propose a causal graph sampler that ranks a node's neighbors based on localized random-walk estimations rather than uniform expansion. By regularizing against high-degree hubs and strictly enforcing temporal constraints, this approach extracts highly relevant, densely connected local subgraphs, which improves local message passing.
  \item \textbf{Task-Optimized Global Memory via Differentiable Codebooks:} We introduce dual memory banks that capture recurring tabular and structural patterns. Using branch-specific cross-attention, the model assigns each seed node a soft membership to these global archetypes, effectively generalizing across disconnected seeds that share similar features or local topology profiles.
  \item \textbf{Asymmetric Cross-Attention for Distant Relational Context:} To capture long-range temporal and collaborative dynamics, we design two specialized global branches. A Perceiver-style bottleneck compresses sprawling sequences of distant neighbors into fixed-capacity latents, enabling efficient macro-level reasoning across structurally distant graph nodes.
\end{enumerate}

\section{Methodology}
\label{sec:method}

In this section, we introduce \textsc{Quartet} (QUad-branch cross-Attention and Random-walk Traces for Enhancing Transformers on relational graphs), a general-purpose Transformer architecture designed for representation learning on heterogeneous temporal relational graphs.
By seamlessly integrating fine-grained, local structural dynamics with global, structurally distant graph contexts, \textsc{Quartet} provides a highly expressive framework for complex relational data.

\begin{figure*}[t]
\centering
\includegraphics[width=\textwidth]{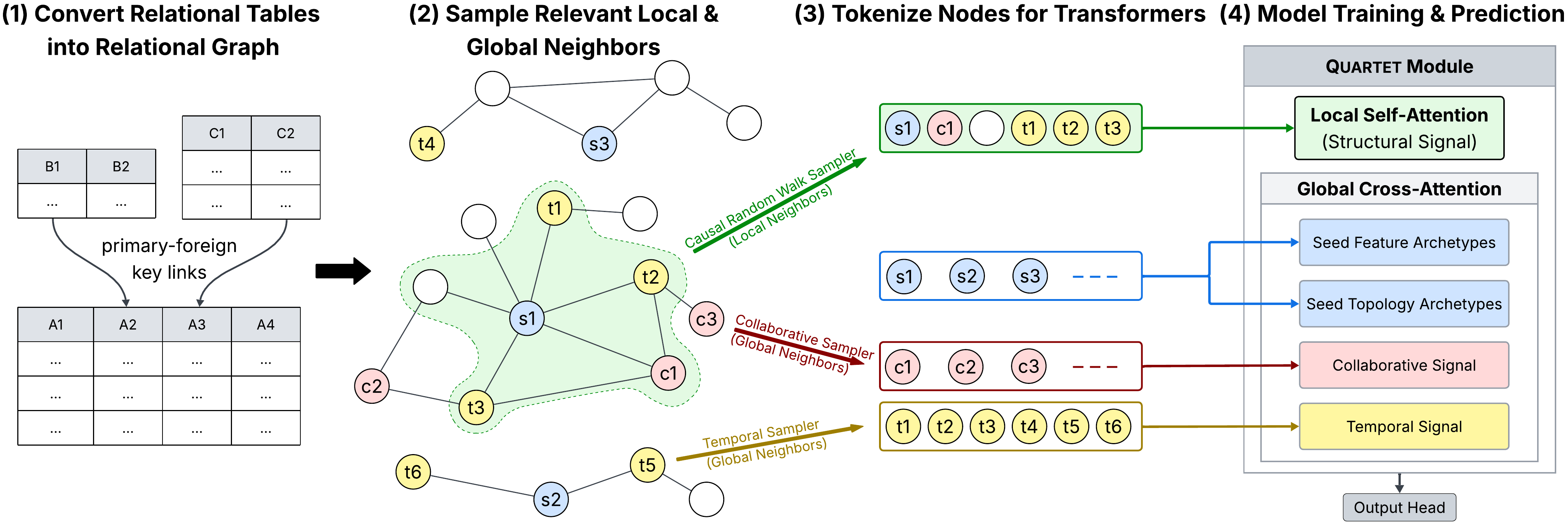}
\caption{Overview of the \textsc{Quartet} pipeline. \textbf{(1)} Relational tables are converted into a heterogeneous temporal graph (\S\ref{sec:formulation}). \textbf{(2)} Three causal samplers extract complementary context: CRW draws a dense \emph{local} subgraph, while temporal and collaborative samplers gather \emph{global} node sequences (\S\ref{sec:sampling}). \textbf{(3)} Nodes are tokenized into fixed-length sequences (\S\ref{sec:tokenization}). \textbf{(4)} Local subgraphs feed into self-attention, while global sequences pass through a quad-branch cross-attention module before final fusion (\S\ref{sec:global}).}
\label{fig:workflow}
\end{figure*}

Figure~\ref{fig:workflow} gives an end-to-end overview of the \textsc{Quartet} pipeline (graph construction, causal local and global sampling, tokenization, and attention-based prediction), while Figure~\ref{fig:arch} details the model architecture.
This architecture is driven by three core components:
(1) \textbf{a Causal Random-Walk (CRW)} subgraph sampler, which constructs a localized subgraph by ranking neighbors according to their recency-truncated Personalized PageRank (PPR)~\citep{page1999pagerank,klicpera2019predict,bojchevski2020pprgo,leskovec2018pinsage} relative to the seed---accumulated over short restart walks so that only the most topologically relevant neighbors are retained;
(2) \textbf{a Local Module}, which processes this sampled subgraph with full self-attention to capture fine-grained structural and temporal dependencies;
(3) \textbf{a Global Module}, which captures macro-level graph dynamics via a quad-branch design.
Within this module, the feature and topology branches extract global archetypes for the seed nodes, while the temporal and collaborative branches identify similarities across structurally distant nodes.

We structure this section as follows: \S\ref{sec:formulation} details graph construction, \S\ref{sec:sampling} introduces the causal samplers, \S\ref{sec:tokenization} covers tokenization, \S\ref{sec:local} and \S\ref{sec:global} present the Local and Global Attention Modules, and \S\ref{sec:fusion} outlines their fusion.

\subsection{Relational Entity Graphs (REG)}
\label{sec:formulation}
To enable end-to-end learning without manual feature engineering, we convert relational databases into Relational Entity Graphs (REGs), the first stage of the pipeline in Figure~\ref{fig:workflow}.
Formally, we model a REG as a heterogeneous temporal graph $G = (V, E, \phi, \psi, \tau)$, where $V$ is the set of nodes (entities) and $E \subseteq V \times V$ is the set of edges (primary-foreign key relationships).
The mapping functions $\phi : V \to \mathcal{T}$ and $\psi : E \to \mathcal{R}$ assign nodes and edges to their respective source tables (node types) and relation types, where $\mathcal{T}$ is the set of node types (with $|\mathcal{T}|$ their number) and $\mathcal{R}$ is the set of relation types.
Finally, $\tau : V \cup E \to \mathbb{R}$ assigns timestamps to temporal entities, treating atemporal nodes as universally available and inducing edge timestamps as the maximum of their endpoints, $\tau(e) = \max\{\tau(u), \tau(v)\}$ for $e = (u, v) \in E$.

Given a seed node $v_i \in V$ and a prediction time $\tseed$, the objective is to produce an embedding of $v_i$ strictly conditioned on information available at or before $\tseed$, thereby enforcing a strict causal constraint that prevents temporal leakage.
We enforce this constraint uniformly across the three samplers of \S\ref{sec:sampling}: every edge traversed during sampling satisfies $\tau(e)\le\tseed$.

\subsection{Causal Samplers for Local and Global Context}
\label{sec:sampling}
\textsc{Quartet} utilizes three seed-rooted causal samplers: CRW, temporal, and collaborative [Figure~\ref{fig:workflow}, stage (2)].
The CRW sampler extracts the causal local subgraph via random walk with restart and also provides the structural sketch for the topological codebook branch.
In parallel, the temporal and collaborative samplers generate the fixed-length global neighbor sequences required by their respective Perceiver cross-attention branches.
Finally, to prevent runtime bottlenecks, all local and global subgraphs are generated once per seed and cached into a memory-mapped file, eliminating the need for graph traversals during training.

\subsubsection{Causal Random Walk (CRW) Local Subgraph Sampler}
\label{sec:ctrwr}
Most standard subgraph extractors fill a $K$-token context by uniform neighborhood sampling or rigid per-type budgeting~\citep{hamilton2017graphsage,zeng2020graphsaint,hu2020hgt}.
On graphs with high-degree hubs these are both computationally prohibitive (scaling with hub degree) and structurally blind (treating all neighbors equally).
Consequently, they exhaust the token budget on random, disconnected slices of massive hubs, yielding sparse subgraphs that severely bottleneck downstream message passing.

To address this limitation, our CRW sampler performs short restart walks from the seed node, ranking neighbors based on a truncated Monte-Carlo estimate of their causal PPR ~\citep{page1999pagerank,klicpera2019predict,bojchevski2020pprgo,leskovec2018pinsage}.
At each step, the walk either teleports back to the seed with probability $p$, or transitions to one of up to $M$ valid causal neighbors of the current node.
By default, we restrict this transition to the $M=50$ most recent neighbors, thereby injecting a strong temporal prior into the sampling process. 
Accumulating visit frequencies across $W$ independent walks of length $L$ yields an empirical estimate of the seed's causal PPR, which converges to the exact distribution as $M, W \to \infty$ (Appendix~\ref{app:ctrwr}).

Crucially, this $M$-truncation serves as a structural regularizer. By capping the per-step branching factor, it bounds the sampling complexity independently of the maximum hub degree and prevents massive hubs from acting as probability sinks. Consequently, the highest-ranked nodes, sharing dense structural pathways with the seed, induce a highly connected and relevance-weighted local subgraph. Detailed derivations of the walk kernel, PPR equivalence, subgraph connectivity, and sparse-neighborhood fallback policies are provided in Appendix~\ref{app:ctrwr}. The exact per-seed procedure is detailed in Algorithm~\ref{alg:rwr} (Appendix~\ref{app:samplers}), with a full computational cost analysis in Appendix~\ref{sec:complexity-samplers}.

\subsubsection{Temporal Global Subgraph Sampler}
\label{sec:temporal-sampler}
While the temporal branch requires a global view of recent graph activities, naively selecting the top-$K$ most recent nodes causes high-frequency tables (e.g., transactions) to disproportionately bias the sample, thereby starving slower-moving dimension tables. 
To address this, the temporal sampler extracts the most recent causal nodes \emph{per node type} under a uniform type-specific budget $B=K_{\mathrm{temp}}/|\mathcal{T}|$, strictly retaining nodes within a temporal lookback window $[\tseed-\Delta t,\tseed]$. 
By executing a single backward-in-time scan, which halts once the time window is exceeded or all per-type budgets are met, we obtain a fixed-length, type-balanced sequence of $K_{\mathrm{temp}}$ IDs (Algorithm~\ref{alg:temporal}, Appendix~\ref{app:samplers}). We set $K_{\mathrm{temp}}=300$ and $\Delta t=365$ days.

\subsubsection{Collaborative Neighbor Sampler}
\label{sec:collab-sampler}
Complementary to the proximity- and recency-biased samplers, the collaborative branch captures long-horizon ``lookalike'' peers---nodes that exhibit substantial neighborhood overlap with the seed across its full causal history, independent of recency. 
This affinity is estimated using a two-hop random probe. By first sampling $M_1$ of the seed's causal neighbors and subsequently drawing $M_2$ peers from each, a peer's accumulated hit count serves as a Monte Carlo estimate of shared intermediaries (representing the numerator of a Jaccard similarity coefficient). 
Peers corresponding to the seed, its direct causal neighbors, or previously sampled entities are discarded. The top $K_{\mathrm{collab}}$ remaining peers by hit count are then considered as collaborative neighbors (Algorithm~\ref{alg:collab} in Appendix~\ref{app:samplers}; derivation in Appendix~\ref{app:collab}). We set $M_1=M_2=50$ and $K_{\mathrm{collab}}=300$.

\begin{figure*}[t]
\centering
\includegraphics[width=\textwidth]{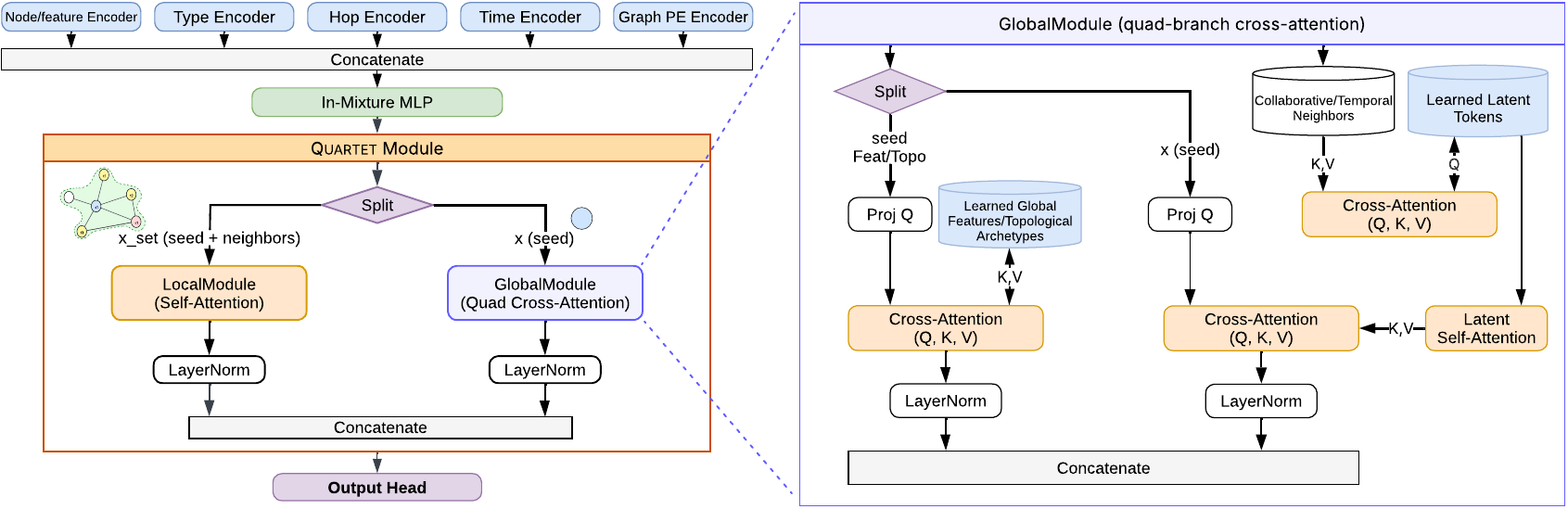}
\caption{The \textsc{Quartet} model. Local tokens fuse five encodings (feature, type, hop, time, positional) via an MLP. Then, the full neighbor set is fed into the \emph{Local Module} (self-attention), while the seed token queries the \emph{Global Module}. This global module cross-attends the seed over learned feature and topological codebooks (\S\ref{sec:codebooks}) and over Perceiver latents compressing temporal and collaborative neighbors (\S\ref{sec:xatt}). Finally, local and global embeddings are concatenated for downstream prediction.}
\label{fig:arch}
\end{figure*}

\subsection{Graph Tokenization}
\label{sec:tokenization}
With the local subgraph sampled, we tokenize it for Transformer consumption: for each seed $v_i$ the context is a fixed-size set of $K$ tokens---the seed and the $K-1$ neighbors ranked by CRW visit frequency (\S\ref{sec:ctrwr}).
Rather than compressing all structural information into a single positional encoding, we decompose the token representation to explicitly model the distinct characteristics of relational data.
Each token $v_j$ is a 5-tuple fused into a single embedding (Figure~\ref{fig:arch}). 
Components include: a PyTorch Frame~\citep{hu2024pytorchframe} feature embedding $h_{\mathrm{feat}}(v_j)$; one-hot encodings for node type $\phi(v_j)$ and hop distance $p(v_i,v_j)$; a linear projection of relative time $\tau(v_j)-\tau(v_i)$; and a Graph Isomorphism Network (GIN)-based positional encoding~\citep{dwivedi2022lspe,lim2023signnet} over the sampled subgraph using stochastic random features~\citep{dwivedi2026relgt}.
These five components are concatenated and mixed via $W_{\mathrm{mix}}\in\mathbb{R}^{d\times 5d}$ to form $h_{\mathrm{token}}(v_j)$ (exact formulas in Appendix~\ref{app:token-encoders}). 
The resulting length-$K$ sequence is then processed by the Local Attention Module (\S\ref{sec:local}).

\subsection{Local Attention Module}
\label{sec:local}
We process the tokenized subgraph using an $L$-layer standard Transformer, applying all-pair self-attention~\citep{vaswani2017attention} across all $K$ tokens of the CRW-sampled subgraph. 
This unconstrained all-pair attention circumvents the traditional GNN bottleneck of localized message passing along connected edges, directly capturing long-range dependencies. 
The final local representation is computed as:
\begin{equation}
    h_{\mathrm{local}}(v_i) = \mathrm{Pool}(\mathrm{FFN}(\mathrm{SelfAttn}(v_i, \{v_j\}_{j=1}^K))_L)
\end{equation}
where $\mathrm{Pool}$ aggregates the sequence via a learned linear combination. 
The CRW formulation ensures these local self-attention layers operate on structurally dense, strictly causal neighborhoods, maximizing the token budget.

\subsection{Global Attention Module}
\label{sec:global}

The Global Attention Module captures long-range relational context beyond the localized subgraph's receptive field—such as structurally disconnected seed archetypes, concurrently active entities, and behaviorally similar peers located multiple hops away.
Architecturally, it consists of four seed-conditioned branches grouped into two functional families with complementary structural goals.
The first family consists of two codebook branches that extract the seed's intrinsic tabular and topological properties by querying learnable archetype matrices. The second family consists of two cross-attention branches that model extrinsic interactions by compressing global temporal and collaborative neighbor sequences via a Perceiver-style bottleneck~\citep{jaegle2021perceiver,lee2019set}.
Unlike prior approaches, all global branches share a unified query-driven bottleneck and are optimized end-to-end. A compact, learnable query set reads over a context tensor, ensuring that the output dimensionality is strictly bounded by a fixed query budget rather than scaling with variable neighborhood sizes. Formally, for a query set $Q \in \mathbb{R}^{d \times n_q}$ and context $C \in \mathbb{R}^{d \times n_c}$ whose columns are tokens, with projection matrices $W_Q, W_K, W_V$ across $H$ heads ($d_k = d / H$), the shared multi-head scaled dot-product cross-attention operator~\citep{vaswani2017attention} applies each projection on the left and normalizes the softmax over the context axis:
\begin{equation}
  \mathrm{CrossAttn}(Q,C) = (W_V C)\,\mathrm{softmax}\!\left(\frac{(W_K C)^\top (W_Q Q)}{\sqrt{d_k}}\right).
  \label{eq:xattn}
\end{equation}

\subsubsection{Feature and Topological Codebook Branches}
\label{sec:codebooks}

Intuitively, the two codebook branches assign each seed a soft membership over a bank of global archetypes: learned prototypes that summarize recurring feature and structural profiles across the entire database.
Because these archetypes are shared globally rather than read off the seed's sampled neighborhood, they let the model relate structurally disconnected nodes that never co-occur in any subgraph.
For example, low- versus high-spend customers are routed to different archetypes according to their tabular attributes (feature codebook) and their local connectivity signature (topological codebook), even when no path in the graph links them.

Each codebook branch queries a learnable archetype matrix, denoted as $E_{\mathrm{struct}}$ or $E_{\mathrm{feat}}$, using the cross-attention operator where the seed token serves as the query [Figure ~\ref{fig:arch}].
To maximize computational efficiency, the key and value projections for both matrices are cached and shared across the batch. For a given branch $b \in \{\mathrm{feat}, \mathrm{struct}\}$, we project a seed query $q_b$ through a two-layer GELU multi-layer perceptron (MLP) and compute the cross-attention against its corresponding archetype matrix $E_b$:
\begin{equation}
  X_b = \mathrm{CrossAttn}\bigl(\mathrm{MLP}_b(q_b),\; E_b\bigr).
  \label{eq:cb}
\end{equation}
The feature query, $q_{\mathrm{feat}} = h_{\mathrm{feat}}(v_i)$, reuses the tabular embedding of the seed generated by the local module. This yields a soft readout of the semantic profile of the seed without introducing additional encoding overhead. Conversely, the structural query summarizes the causal one-hop neighborhood of the seed via a structural sketch $\Fsketch(v_i)$. This sketch concatenates four scalar random-walk statistics with a per-type degree histogram $\mathbf{c}$, formulated as:
\begin{equation}
  \Fsketch(v_i)=\log\!\bigl(1+[\, d_c,\ s,\ \rho,\ \pi^{\max},\ \mathbf{c} \,]\bigr).
  \label{eq:sketch}
\end{equation}
Here $d_c$ is the causal degree (the number of one-hop neighbors with $\tau(e)\le\tseed$), $s$ the temporal span of those neighbors ($t_{\max}-t_{\min}$), $\rho=d_c/s$ their arrival density, and $\pi^{\max}$ the peak causal-PPR mass (the largest CRW visit count), while $\mathbf{c}\in\mathbb{R}^{|\mathcal{T}|}$ is the per-type degree histogram counting causal neighbors of each node type.
The elementwise $\log(1+\cdot)$ compresses these heavy-tailed statistics and ensure empty neighborhoods are mapped to zero, yielding a compact, permutation-invariant fingerprint of the seed's local connectivity: its size, temporal extent, interaction rate, dominant-neighbor concentration, and type mix.
Applying Batch Normalization and an MLP projection to this sketch yields $q_{\mathrm{struct}} = \mathrm{BN}\!\bigl(\Fsketch(v_i)\bigr)$, which provides a soft summary over the global topological archetypes to capture structural homophily.

\subsubsection{Temporal and Collaborative Cross-Attention Branches}
\label{sec:xatt}

To circumvent the quadratic cost of full self-attention over extended neighbor sequences, the temporal and collaborative branches each employ an asymmetric cross-attention bottleneck.
Each track $b \in \{\mathrm{temp}, \mathrm{collab}\}$ (indexed as in \S\ref{sec:codebooks}) instantiates its own learnable latent set $\mathbf{Z}_0^{b} \in \mathbb{R}^{d \times L_{\mathrm{perc}}}$; the two branches are separate bottlenecks and share no latent tokens.
For branch $b$, these latents cross-attend over the branch context $\mathbf{C}_b$---the featurized set of sampled temporal or collaborative neighbors (\S\ref{sec:sampling})---with a residual connection:
\begin{equation}
  \mathbf{Z}_b = \mathbf{Z}_0^{b} + \mathrm{CrossAttn}(\mathbf{Z}_0^{b}, \mathbf{C}_b), \qquad b \in \{\mathrm{temp}, \mathrm{collab}\}.
  \label{eq:cross-attn-gen}
\end{equation}
The temporal context $\mathbf{C}_{\mathrm{temp}}$ is constructed from recently active nodes to isolate transient global phenomena, such as transactional waves, independently of spatial topology.
For instance, in the Formula-1 tasks (driver-top3, driver-dnf), a driver's finishing outcome is shaped by conditions that act on the entire grid at once---such as weather or a safety-car period---signals that live in concurrently active nodes rather than in the driver's own local neighborhood.
In contrast, the collaborative context $\mathbf{C}_{\mathrm{collab}}$ aggregates highly co-occurring, multi-hop peers to distill long-term behavioral affinities and shared archetypes while strictly bypassing recency bias.
For instance, in e-commerce (rel-amazon) or fashion retail (rel-hm), two customers who never transact together may still be strong lookalikes if they repeatedly buy the same products, and their history is highly predictive for repeat-purchase and churn targets.

Rather than pooling the two blocks in isolation, we let them exchange information before readout: the latent blocks are concatenated along the token axis, jointly reasoned over by a shared stack of $L_{\mathrm{sa}}$ pre-LN self-attention layers ($L_{\mathrm{sa}} = 4$), and split back into per-branch halves,
\begin{equation}
  [\,\tilde{\mathbf{Z}}_{\mathrm{collab}} \parallel \tilde{\mathbf{Z}}_{\mathrm{temp}}\,] = \mathrm{SelfAttn}\bigl([\,\mathbf{Z}_{\mathrm{collab}} \parallel \mathbf{Z}_{\mathrm{temp}}\,]\bigr).
  \label{eq:latent-sa}
\end{equation}
This cross-branch reasoning step lets the temporal and collaborative latents attend to one another, so each track is refined in the context of the other rather than in isolation.
Each refined block is then collapsed to a flat vector by a seed-conditioned cross-attention read that reuses the single-token query convention of \S\ref{sec:codebooks}, with the projected seed token $q_{\mathrm{seed}}$ as the query:
\begin{equation}
  h_b = \mathrm{CrossAttn}(q_{\mathrm{seed}}, \tilde{\mathbf{Z}}_b), \qquad b \in \{\mathrm{temp}, \mathrm{collab}\}.
  \label{eq:seed-readout}
\end{equation}
The resulting vectors $h_{\mathrm{temp}}, h_{\mathrm{collab}} \in \mathbb{R}^d$ are forwarded to the final fusion (\S\ref{sec:fusion}).

\subsection{Final Fusion}
\label{sec:fusion}

The outputs of the four global branches ($X_{\mathrm{struct}}, X_{\mathrm{feat}}, h_{\mathrm{collab}}, h_{\mathrm{temp}}$) are normalized via independent LayerNorm ($\mathrm{LN}$) layers, concatenated with the local embedding $h_{\mathrm{local}}$, and linearly projected back to $d$ dimensions:
\begin{equation}
  h_{\mathrm{fused}} = W_{\mathrm{proj}} \bigl[\, h_{\mathrm{local}} \parallel \mathrm{LN}(X_{\mathrm{struct}}) \parallel \mathrm{LN}(X_{\mathrm{feat}}) \parallel \mathrm{LN}(h_{\mathrm{collab}}) \parallel \mathrm{LN}(h_{\mathrm{temp}}) \,\bigr].
  \label{eq:fusion}
\end{equation}
The prediction head subsequently computes the target output as $y_{v_i} = \mathrm{head}(\mathrm{FFN}(h_{\mathrm{fused}}))$. 
By concatenating rather than summing these distinct channels, we preserve their semantic separability and enable $W_{\mathrm{proj}}$ to learn an expressive, per-branch mixing rule tailored for downstream tasks.

\section{Experimental Setup}
\label{sec:setup}

We evaluate our architecture across twelve binary-classification tasks derived from seven \textsc{RelBench} datasets~\citep{gu2026relbench}, which span diverse real-world domains such as Formula-1 racing, event recommendation, clinical trials, online advertising, e-commerce, fashion retail, and Q\&A communities.
We use ROC-AUC as the primary evaluation metric and report test ROC-AUC for the model selected on the validation split.
All models are trained end-to-end using the Adam optimizer.
To accommodate varying computational demands, small tasks (<1M training nodes) are trained on a single A100 GPU, whereas large tasks (>1M training nodes) utilize four A100 GPUs.
To eliminate runtime traversal overhead, every subgraph is pre-computed and cached offline prior to training.
Following a similar hyperparameter tuning strategy to that described in the RelGT paper, we adjust the number of local layers ($L_{\mathrm{local}}$), dropout, and weight decay on a per-task basis.
Specifically, we search $L_{\mathrm{local}} \in \{1, 4, 8\}$ for small datasets, while compute constraints require us to fix $L_{\mathrm{local}} = 4$ for large datasets (detailed in Appendix~\ref{app:hp}).
Furthermore, the quad-branch capacity---specifically the Perceiver latent budget $L_{\mathrm{perc}}$ and codebook size $R_{\mathrm{struct}}=R_{\mathrm{feat}}$---is configured per task (see Appendix~\ref{app:hp-capacity}), and the CRW local sampler alongside the collaborative and temporal branches each uses token budget of $K=300$.
Full configuration grids, sampler settings, and infrastructure details are provided in Appendix~\ref{sec:impl}.
While we retain the twelve classification tasks established in \textsc{RelBench} v1 to align with prior work, all data processing and feature tokenization are executed through the updated, more rigorous \textsc{RelBench} v2 pipeline (using the most up-to-date datasets)~\citep{gu2026relbench}.
To ensure a fair evaluation across all models, we reproduced both graph-transformer baselines (HGT and RelGT) from scratch under this shared v2 pipeline, utilizing the original authors' reported hyperparameters.
Finally, to account for architectural stochasticity, all experiments for both the baselines and \textsc{Quartet} are averaged across $n=4$ random seeds.

\section{Results and Discussion}
\label{sec:results}

We evaluate \textsc{Quartet} in three stages.
First, we isolate the performance improvements of the causal random-walk sampler (CRW) by swapping out the default 2-hop BFS sampler with CRW in the current state-of-the-art RelGT workflow (\S\ref{sec:results-rwr}).
Second, we benchmark the complete \textsc{Quartet} model against previous SOTA baselines on the \textsc{RelBench} v1 classification tasks (\S\ref{sec:results-full}).
Finally, we conduct a leave-one-out ablation study to determine the specific contributions of the local module, the CRW sampler, and each of the four global branches (\S\ref{sec:results-ablation}).

\subsection{Causal Random-Walk Sampling Allows Better Graph Representation Learning with Fewer Local Attention Layers}
\label{sec:results-rwr}

In this section, we test the hypothesis that ranking a seed's neighbors by causal random walk, yields a more natural, better-connected subgraph than uniform neighbor selection, and that this structurally relevant local context lets the encoder learn substantially stronger graph representations.
To isolate the sampler from every other design choice, we hold the base RelGT architecture~\citep{dwivedi2026relgt} fixed and vary only the neighbor sampler.
The baseline relies on RelGT's native \emph{naive} sampler, which draws neighbors uniformly at random from the seed's one- and two-hop (BFS) neighborhood. Our system substitutes this with the PPR-ranked CRW sampler. To ensure a fair comparison, both configurations utilize the same local attention architecture and follow an identical restricted hyperparameter tuning protocol (optimizing for layers, dropout, and weight decay), as detailed in Appendix~\ref{app:hp}.
We report the test ROC-AUC for all classification tasks in \textsc{RelBench} v1.

\begin{figure}[t]
\centering
\includegraphics[width=\textwidth]{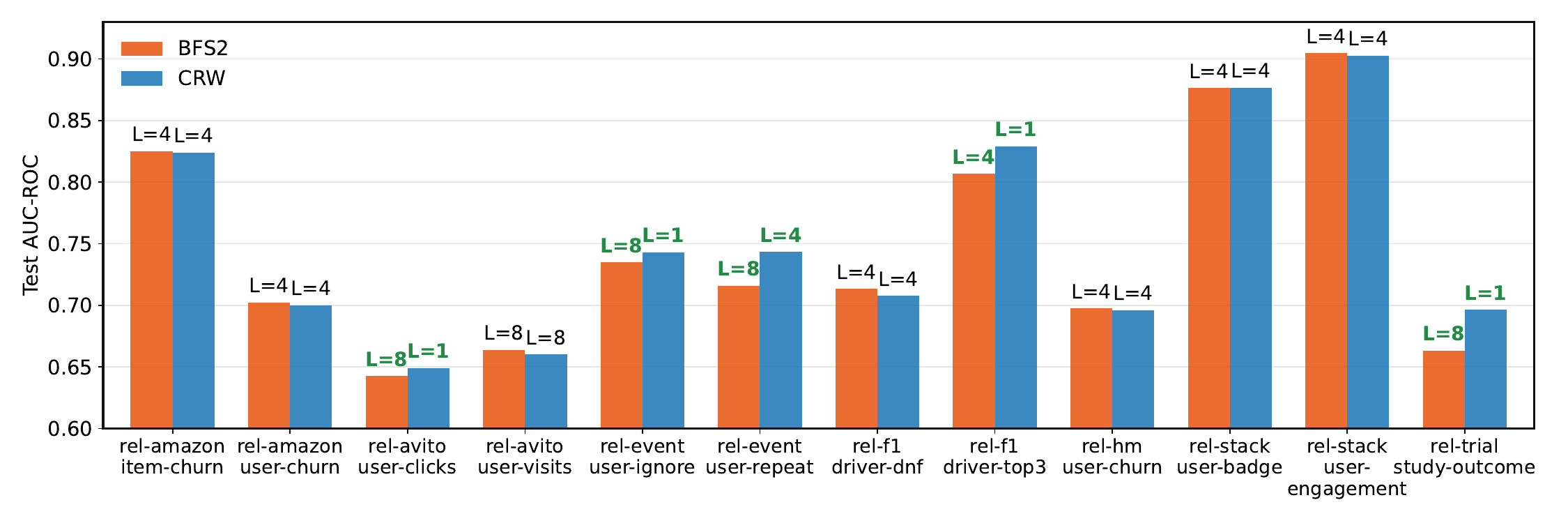}
\caption{Per-task test ROC-AUC on the \textsc{RelBench} classification tasks for the PPR-ranked CRW sampler (\emph{CRW}, blue) against RelGT's naive uniform one-/two-hop sampler (\emph{BFS2}, orange) on a fixed RelGT architecture.
Each bar is annotated with the selected number of local layers $L_{\mathrm{local}}$.}
\label{fig:rwr-vs-bfs2}
\end{figure}

Figure~\ref{fig:rwr-vs-bfs2} demonstrates that the CRW sampler matches or outperforms the RelGT baseline while requiring fewer local self-attention layers (annotated in green).
By concentrating a seed's most relevant neighbors into a compact, well-connected subgraph, CRW enables the local module to learn stronger representations with less depth.
Ultimately, this sampler upgrade delivers both improved accuracy and a more efficient local module.
A detailed subgraph connectivity analysis further confirming that CRW extracts denser, more connected subgraphs than the BFS2 baseline is provided in Appendix~\ref{app:crw-bfs-structure}.
For the experiments in the next section, we reused the result of hyperparameter search from this section for number of local layers $L_{\mathrm{local}}$, dropout, and weight decay (see Appendix~\ref{app:hp}).

\subsection{\textsc{Quartet} Outperforms RelGT on Most Classification Tasks}
\label{sec:results-full}

Table~\ref{tab:baselines} compares \textsc{Quartet} against leading graph transformers (HGT~\citep{hu2020hgt} and RelGT~\citep{dwivedi2026relgt}) on the \textsc{RelBench} v1 classification tasks. Compared directly to its base architecture, \textsc{Quartet} improves on or matches RelGT on eight of twelve tasks. RelGT primarily maintains an advantage on large, stationary tasks where dense local topology dominates the predictive signal, rendering additional global branches redundant. Overall, \textsc{Quartet} is competitive with the strongest graph-transformer baselines and lifts its RelGT starting point on the classification tasks; the following ablation section (\S\ref{sec:results-ablation}) isolates where these predictive gains originate.

\begin{table*}[t]
\centering
\scriptsize
\caption{Test ROC-AUC (mean\,$\pm$\,std) on the \textsc{RelBench} classification suite. $\Delta\%$ compares \textsc{Quartet} with RelGT (positive $=$ gain); \textbf{bold} marks the best mean in each row.}
\label{tab:baselines}
\resizebox{\textwidth}{!}{%
\begin{tabular}{llcccc}
\toprule
Dataset & Task & HGT & RelGT & \textsc{Quartet} & $\Delta\%$ vs.\ RelGT \\
\midrule
rel-amazon & item-churn     & 0.7758 {\scriptsize $\pm$ 0.0075} & 0.8249 {\scriptsize $\pm$ 0.0005} & \textbf{0.8262} {\scriptsize $\pm$ 0.0010} & \cellcolor{relgtblue!25}$+0.2$ \\
           & user-churn     & 0.6626 {\scriptsize $\pm$ 0.0041} & \textbf{0.7024} {\scriptsize $\pm$ 0.0015} & 0.7001 {\scriptsize $\pm$ 0.0016} & \cellcolor{red!25}$-0.3$ \\
\midrule
rel-avito & user-clicks     & \textbf{0.6501} {\scriptsize $\pm$ 0.0104} & 0.6425 {\scriptsize $\pm$ 0.0159} & 0.6481 {\scriptsize $\pm$ 0.0122} & \cellcolor{relgtblue!25}$+0.9$ \\
          & user-visits     & 0.6410 {\scriptsize $\pm$ 0.0068} & \textbf{0.6639} {\scriptsize $\pm$ 0.0020} & 0.6508 {\scriptsize $\pm$ 0.0050} & \cellcolor{red!25}$-2.0$ \\
\midrule
rel-event & user-ignore     & 0.7256 {\scriptsize $\pm$ 0.0080} & 0.7348 {\scriptsize $\pm$ 0.0183} & \textbf{0.7392} {\scriptsize $\pm$ 0.0035} & \cellcolor{relgtblue!25}$+0.6$ \\
          & user-repeat     & 0.7539 {\scriptsize $\pm$ 0.0157} & 0.7160 {\scriptsize $\pm$ 0.0316} & \textbf{0.7563} {\scriptsize $\pm$ 0.0175} & \cellcolor{relgtblue!25}$+5.6$ \\
\midrule
rel-f1 & driver-dnf         & 0.6595 {\scriptsize $\pm$ 0.0627} & \textbf{0.7133} {\scriptsize $\pm$ 0.0416} & 0.7044 {\scriptsize $\pm$ 0.0180} & \cellcolor{red!25}$-1.2$ \\
       & driver-top3        & 0.7384 {\scriptsize $\pm$ 0.0574} & 0.8067 {\scriptsize $\pm$ 0.0402} & \textbf{0.8517} {\scriptsize $\pm$ 0.0156} & \cellcolor{relgtblue!25}$+5.6$ \\
\midrule
rel-hm & user-churn         & 0.6746 {\scriptsize $\pm$ 0.0036} & 0.6975 {\scriptsize $\pm$ 0.0016} & \textbf{0.6976} {\scriptsize $\pm$ 0.0011} & \cellcolor{relgtblue!25}$+0.01$ \\
\midrule
rel-stack & user-badge      & 0.8688 {\scriptsize $\pm$ 0.0025} & \textbf{0.8764} {\scriptsize $\pm$ 0.0016} & 0.8528 {\scriptsize $\pm$ 0.0094} & \cellcolor{red!25}$-2.7$ \\
          & user-engagement & 0.8899 {\scriptsize $\pm$ 0.0018} & 0.9047 {\scriptsize $\pm$ 0.0009} & \textbf{0.9063} {\scriptsize $\pm$ 0.0010} & \cellcolor{relgtblue!25}$+0.2$ \\
\midrule
rel-trial & study-outcome   & 0.6345 {\scriptsize $\pm$ 0.0251} & 0.6632 {\scriptsize $\pm$ 0.0177} & \textbf{0.7059} {\scriptsize $\pm$ 0.0044} & \cellcolor{relgtblue!25}$+6.4$ \\
\bottomrule
\end{tabular}%
}
\end{table*}

\subsection{CRW Enriches Local Context While Global Branches Add Complementary Gains}
\label{sec:results-ablation}

To rigorously isolate the individual contributions of each architectural component, we conduct a leave-one-out ablation study (Table~\ref{tab:ablation}).
Starting from the full \textsc{Quartet} model, we systematically remove a single component and record the relative changes in test ROC-AUC.
A performance degradation indicates that the ablated component provided essential predictive signals that the remaining architecture could not independently recover.
Appendix Table~\ref{tab:ablation-full} reports the full mean\,$\pm$\,std values underlying these relative changes, with statistically significant drops highlighted.

\begin{table*}[t]
\centering
\footnotesize
\setlength{\tabcolsep}{2.5pt}
\caption{Leave-one-out ablation of \textsc{Quartet}: entries are the relative change in test ROC-AUC (\%) vs.\ the full model. Cell color encodes the change (\textcolor{red!70}{red}~$=$~decrease, \textcolor{relgtblue!70}{blue}~$=$~increase; darker is larger in magnitude, with the diverging scale saturating at $-10\%$ and $+1\%$). Columns are grouped into attention variants (No Local, No Perceiver), sampler mechanisms (No CRW, No Hub Trunc., No Restart), and global branches (No Temp., No Collab., No Feat., No Struct.). For each task, the largest drop among the four global-branch columns is marked with a \fbox{solid border}.}
\label{tab:ablation}
\begin{tabular}{l l c c @{\hskip 1em} c c c @{\hskip 1em} c c c c}
\toprule
\multirow{2}{*}{Dataset} & \multirow{2}{*}{Task} & No & No & No & No Hub & No & No & No & No & No \\
 & & Local & Perc. & CRW & Trunc. & Restart & Temp. & Collab. & Feat. & Struct. \\
\midrule
rel-amazon & item-churn     & \cellcolor{red!69}$-9.8$  & \cellcolor{red!1}$-0.2$  & \cellcolor{red!1}$-0.2$  & \cellcolor{red!1}$-0.2$  & \cellcolor{red!1}$-0.1$  & \cellcolor{red!1}$\boxed{-0.2}$   & \cellcolor{red!1}$-0.1$  & \cellcolor{red!1}$-0.1$  & \cellcolor{red!1}$-0.1$  \\
           & user-churn     & \cellcolor{red!37}$-5.3$  & \cellcolor{relgtblue!7}$+0.1$ & \cellcolor{red!8}$-1.1$  & \cellcolor{red!2}$-0.3$  & \cellcolor{red!5}$-0.7$  & \cellcolor{red!1}$-0.1$   & \cellcolor{relgtblue!7}$+0.1$ & \cellcolor{red!3}$-0.4$  & \cellcolor{red!4}$\boxed{-0.6}$  \\
\midrule
rel-avito & user-clicks     & \cellcolor{red!57}$-8.2$  & \cellcolor{red!15}$-2.1$  & \cellcolor{relgtblue!56}$+0.8$ & \cellcolor{relgtblue!42}$+0.6$ & \cellcolor{red!70}$-10.1$ & \cellcolor{red!8}$-1.1$   & \cellcolor{red!28}$-4.0$ & \cellcolor{red!40}$\boxed{-5.7}$ & \cellcolor{red!13}$-1.8$ \\
          & user-visits     & \cellcolor{red!11}$-1.5$  & \cellcolor{red!0}$-0.01$  & \cellcolor{red!8}$-1.2$  & \cellcolor{relgtblue!0}$+0.01$  & \cellcolor{red!6}$-0.9$  & \cellcolor{red!8}$-1.1$   & \cellcolor{red!23}$\boxed{-3.3}$ & \cellcolor{red!6}$-0.8$  & \cellcolor{red!19}$-2.7$ \\
\midrule
rel-event & user-ignore     & \cellcolor{relgtblue!70}$+4.7$ & \cellcolor{red!6}$-0.8$  & \cellcolor{red!6}$-0.8$  & \cellcolor{relgtblue!0}$+0.01$  & \cellcolor{red!1}$-0.2$  & \cellcolor{red!11}$\boxed{-1.6}$  & \cellcolor{red!12}$-1.7$ & \cellcolor{red!9}$-1.3$  & \cellcolor{relgtblue!14}$+0.2$ \\
          & user-repeat     & \cellcolor{red!70}$-25.4$ & \cellcolor{red!13}$-1.8$  & \cellcolor{red!43}$-6.1$ & \cellcolor{red!70}$-10.4$ & \cellcolor{red!16}$-2.3$  & \cellcolor{red!26}$\boxed{-3.7}$  & \cellcolor{red!11}$-1.5$ & \cellcolor{red!19}$-2.7$ & \cellcolor{red!11}$-1.6$ \\
\midrule
rel-f1 & driver-dnf         & \cellcolor{red!30}$-4.3$  & \cellcolor{red!40}$-5.7$  & \cellcolor{red!3}$-0.4$  & \cellcolor{relgtblue!56}$+0.8$ & \cellcolor{relgtblue!14}$+0.2$ & \cellcolor{red!33}$\boxed{-4.7}$  & \cellcolor{red!20}$-2.8$ & \cellcolor{red!25}$-3.6$ & \cellcolor{red!24}$-3.4$ \\
       & driver-top3        & \cellcolor{red!70}$-16.9$ & \cellcolor{red!49}$-7.0$  & \cellcolor{red!25}$-3.5$ & \cellcolor{red!62}$-8.9$  & \cellcolor{red!28}$-4.0$  & \cellcolor{red!32}$\boxed{-4.5}$  & \cellcolor{red!26}$-3.7$ & \cellcolor{red!9}$-1.3$  & \cellcolor{red!22}$-3.2$ \\
\midrule
rel-hm & user-churn         & \cellcolor{red!23}$-3.3$  & \cellcolor{red!1}$-0.1$  & \cellcolor{red!1}$-0.2$  & \cellcolor{relgtblue!14}$+0.2$ & \cellcolor{red!1}$-0.2$  & \cellcolor{red!3}$\boxed{-0.4}$   & \cellcolor{red!1}$-0.1$  & \cellcolor{red!2}$-0.3$  & \cellcolor{red!2}$-0.3$  \\
\midrule
rel-stack & user-badge      & \cellcolor{red!29}$-4.1$  & \cellcolor{red!55}$-7.8$  & \cellcolor{red!15}$-2.2$ & \cellcolor{red!22}$-3.1$  & \cellcolor{red!34}$-4.8$  & \cellcolor{red!30}$-4.3$  & \cellcolor{red!31}$\boxed{-4.4}$ & \cellcolor{red!21}$-3.0$ & \cellcolor{red!18}$-2.5$ \\
          & user-engagement & \cellcolor{red!32}$-4.6$  & \cellcolor{relgtblue!0}$+0.01$ & \cellcolor{red!1}$-0.1$  & \cellcolor{red!0}$-0.01$  & \cellcolor{red!1}$-0.1$  & \cellcolor{red!1}$-0.1$   & \cellcolor{red!1}$-0.2$  & \cellcolor{red!2}$\boxed{-0.3}$  & \cellcolor{red!1}$-0.2$ \\
\midrule
rel-trial & study-outcome   & \cellcolor{red!13}$-1.9$  & \cellcolor{red!4}$-0.6$  & \cellcolor{red!25}$-3.6$ & \cellcolor{red!2}$-0.3$  & \cellcolor{red!13}$-1.8$  & \cellcolor{red!13}$-1.8$  & \cellcolor{red!1}$-0.2$  & \cellcolor{red!8}$-1.2$  & \cellcolor{red!17}$\boxed{-2.4}$ \\
\midrule
\multicolumn{2}{l}{\textbf{Average}} & \cellcolor{red!47}$-6.7$  & \cellcolor{red!15}$-2.2$  & \cellcolor{red!11}$-1.6$ & \cellcolor{red!13}$-1.8$ & \cellcolor{red!15}$-2.1$ & \cellcolor{red!14}$-2.0$  & \cellcolor{red!13}$-1.8$ & \cellcolor{red!12}$-1.7$ & \cellcolor{red!11}$-1.5$ \\
\bottomrule
\end{tabular}
\end{table*}

\paragraph{Architecture ablation.}
Removing the local module and CRW sampler ("No Local" configuration) yields the steepest average performance drop ($6.7\%$), confirming local topology as the primary predictive signal. However, this is task-dependent; for example, removing local context actively improves performance on rel-event user-ignore.
We also probe the design of the global module by replacing the Perceiver bottleneck with direct full self-attention between the seed and its global neighbor sequences (``No Perceiver''). This variant degrades performance by $2.2\%$ on average, confirming that the latent bottleneck improves representation quality rather than serving as a mere efficiency shortcut.

\paragraph{Sampler ablation.}
Substituting CRW with a naive BFS2 expansion (``No CRW'') degrades performance by $1.6\%$ on average. As expected, the tasks that suffer the steepest performance drops largely align with the tasks that saw the largest CRW-driven gains in our standalone sampler study (\S\ref{sec:results-rwr}), confirming that the local encoder relies heavily on CRW's structurally dense topology.
To further dissect the CRW sampler, we ablate its two key mechanisms: hub-degree truncation (``No Hub Trunc.'') and the restart probability (``No Restart'').
Disabling hub truncation degrades performance by $1.8\%$ on average.
This drop is especially severe in the rel-event user-repeat and rel-f1 driver-top3 tasks, where high-degree hub nodes will dominate walk distributions if left unchecked.
Removing the restart mechanism yields a comparable average drop of $2.1\%$, most acutely on rel-avito user-clicks and rel-stack user-badge, indicating that restart-driven locality is important when the predictive signal concentrates near the seed node.
Because these $1.8\%$ and $2.1\%$ drops exceed the $1.6\%$ penalty of dropping CRW entirely, this confirms both mechanisms are essential; without them, CRW actually underperforms a naive BFS2 sampler.

\paragraph{Quad-branch ablation.}
The four global branches (Temporal, Collaborative, Feature, and Structural) act as complementary boosters, yielding average gains of $2.0\%$, $1.8\%$, $1.7\%$, and $1.5\%$, respectively.
No single branch is globally redundant, as the dominant component shifts dynamically across tasks (solid borders, Table~\ref{tab:ablation}), though the temporal branch sustains the largest drop on half of the evaluation tasks.

Ultimately, this ablation suggests that \textsc{Quartet}'s empirical strength stems from resolving hub-sensitivity via PPR-ranked sampling and augmenting the local signal with rich, complementary global context.

\section{Conclusion}
\label{sec:conclusion}

In this work, we introduced \textsc{Quartet}, a quad-branch architecture extending Relational Graph Transformer (RelGT) by augmenting local subgraphs with a Causal Random Walk (CRW) sampler and incorporating global context via Perceiver- and codebook-style cross-attention branches.
Evaluated on the \textsc{RelBench} v1 classification benchmark, \textsc{Quartet} achieves competitive performance against strong graph-transformer baselines, matching or outperforming the original RelGT model on eight of twelve tasks and securing top test ROC-AUC on seven.
While RelGT retains an edge on a few large, stationary tasks, our ablations demonstrate the distinct utility of \textsc{Quartet}'s individual components.
Specifically, the CRW sampler ranks neighbors via a causal PPR estimate and serves as a compelling alternative to uniform BFS expansion, markedly improving test performance while requiring fewer attention layers.
Furthermore, leave-one-out experiments confirm that global context complements rather than replaces local topology: removing the local module (and CRW sampler along with it) causes the most severe performance drop as expected, whereas the four global branches supply complementary and task-dependent signals.
Together, these findings demonstrate that \textsc{Quartet} provides a promising framework for enriching supervised RDL models with global relational context while preserving crucial local attention mechanisms.

\section{Future Directions}
While our current evaluation focuses on twelve binary classification tasks, future work will extend QUARTET to encompass regression modeling and pair-wise link prediction. For link prediction, the architecture can be readily adapted by treating candidate entities as independent seed nodes, extracting their enriched graph embeddings through our pipeline, and applying a shallow classifier to evaluate their relational affinity. Similarly, the model can natively support regression tasks by modifying the final prediction head to output continuous target variables. Furthermore, to validate the architecture's scalability and generalizability in highly realistic enterprise resource planning (ERP) environments, we plan to comprehensively benchmark QUARTET on the SALT dataset.


\bibliographystyle{unsrtnat}
\bibliography{sample-base}

\appendix
\clearpage
\raggedbottom
\makeatletter
\setlength{\@fptop}{0pt}
\makeatother

\section{Appendix}

\smallskip
\noindent\textbf{Table of Contents}
\smallskip
\hrule
\smallskip

\noindent
\begin{tabular}{@{}r@{\;\;}l@{}}
  \S\ref{app:model-details} & \textbf{Model Details} \\[2pt]
  & \quad \S\ref{app:samplers}\;\; Neighbor Sampling Algorithms \\
  & \quad \S\ref{app:ctrwr}\;\; CRW Sampler: Walk Kernel, PPR Equivalence, Connectivity, and Edge Cases \\
  & \quad \S\ref{app:collab}\;\; Collaborative Sampler: Two-Hop Probe as a Jaccard-Numerator Estimator \\
  & \quad \S\ref{app:token-encoders}\;\; Token Encoder Details \\[6pt]
  \S\ref{app:complexity} & \textbf{Complexity Analysis} \\[2pt]
  & \quad \S\ref{sec:complexity-samplers}\;\; Subgraph Sampler Time Complexity \\
  & \quad \S\ref{sec:complexity-attention}\;\; Attention Module Time Complexity \\
  & \quad \S\ref{sec:complexity-empirical}\;\; Empirical Runtime, Memory, and Storage Comparison \\[6pt]
  \S\ref{app:experiment-details} & \textbf{Experiment Details} \\[2pt]
  & \quad \S\ref{sec:impl}\;\; Full Implementation and Training Details \\
  & \quad \S\ref{app:hp-strategy}\;\; Hyperparameter Tuning Strategy and Baseline Comparisons \\
  & \quad \S\ref{app:hp}\;\; Per-Task Hyperparameters \\
  & \quad \S\ref{app:hp-capacity}\;\; Quad-Branch Capacity Search \\[6pt]
  \S\ref{app:additional-results} & \textbf{Additional Results} \\[2pt]
  & \quad \S\ref{app:crw-bfs-structure}\;\; Subgraph Connectivity Analysis \\
  & \quad \S\ref{app:baselines-full}\;\; Validation and Test Scores for All Baselines \\
  & \quad \S\ref{app:ablation-full}\;\; Full Ablation Results with Seed Variability \\
  & \quad \S\ref{app:rgp-reproduction}\;\; RGP Baseline Reproduction \\
\end{tabular}

\smallskip
\hrule
\bigskip

\subsection{Model Details}
\label{app:model-details}

\subsubsection{Neighbor Sampling Algorithms}
\label{app:samplers}
This appendix section provides the algorithms for the three seed-rooted causal samplers discussed in \S\ref{sec:sampling}. The procedures are outlined as follows: Algorithm~\ref{alg:rwr} presents the CRW local subgraph sampler (\S\ref{sec:ctrwr}); Algorithm~\ref{alg:temporal} outlines the temporal sampler's single backward-in-time scan (\S\ref{sec:temporal-sampler}); and Algorithm~\ref{alg:collab} illustrates the two-hop random probe used by the collaborative sampler (\S\ref{sec:collab-sampler}).
\begin{algorithm}[H]
\caption{CRW local subgraph sampling for one seed.}
\label{alg:rwr}
\begin{algorithmic}[1]
\Require Graph $G$; seed $\seed=(v_i,\tseed)$; walks $W$, length $L$, restart $p$, truncation $M$; budget $K$.
\Ensure Sampled neighbor list ($\mathtt{neighbors}$), induced sub-adjacency ($A_S$), hop labels ($h$).
\State $\mathrm{visit}[\cdot] \gets 0$;\quad $\hwalk[\cdot] \gets \infty$
\For{$w = 1$ \textbf{to} $W$}
  \State $u \gets v_i$,\quad $k \gets 0$
  \For{$\ell = 1$ \textbf{to} $L$}
    \If{$\mathrm{Uniform}(0,1) < p$ \textbf{ or } $\Nval_M(u;\seed)=\varnothing$}
      \State $u \gets v_i$,\quad $k \gets 0$
    \Else
      \State $u \sim \mathrm{Uniform}(\Nval_M(u;\seed))$;\quad $k \gets k+1$
      \State $\mathrm{visit}[u] \mathrel{+}= 1$;\quad $\hwalk[u] \gets \min(\hwalk[u], k)$
    \EndIf
  \EndFor
\EndFor
\State $\mathtt{neighbors} \gets [\,v_i\,] \,\|\,$ (top $K-1$ visited non-seed neighbors)
\State $A_S \gets$ causal edges of $G$ induced on $\mathtt{neighbors}$
\State $\hbfs \gets$ BFS hop distance from $v_i$ within $A_S$
\State $h[v] \gets \min\bigl(\hbfs[v],\ \hwalk[v]\bigr)$ for each $v \in \mathtt{neighbors}$;\quad $h[v_i] \gets 0$
\State \Return $\mathtt{neighbors},\ A_S,\ h$
\end{algorithmic}
\end{algorithm}

\begin{algorithm}[H]
\caption{Temporal neighbor sampling for one seed.}
\label{alg:temporal}
\begin{algorithmic}[1]
\Require Seed $\seed=(v_i,\tseed)$; per-type budget $B$; window $\Delta t$; types $\mathcal{T}$.
\Ensure Fixed-length neighbor list $N_{\mathrm{temp}}$ of size $B\cdot|\mathcal{T}|$.
\State $c_a \gets 0$ for all $a\in\mathcal{T}$;\quad $N_{\mathrm{temp}} \gets [\,]$
\For{all nodes $(v, a, t)$ in $G$ with $t \le \tseed$, in \textbf{decreasing} $t$}
  \If{$t < \tseed - \Delta t$ \textbf{ or } $c_a = B$ for all $a\in\mathcal{T}$}
    \State \textbf{break}
  \EndIf
  \If{$c_a < B$ \textbf{ and } $v \notin N_{\mathrm{temp}}$}
    \State append $v$ to $N_{\mathrm{temp}}$;\quad $c_a \gets c_a + 1$
  \EndIf
\EndFor
\State pad $N_{\mathrm{temp}}$ to size $B\cdot|\mathcal{T}|$ via uniform fallback (see Edge Cases)
\State \Return $N_{\mathrm{temp}}$
\end{algorithmic}
\end{algorithm}

\begin{algorithm}[H]
\caption{Collaborative neighbor sampling for one seed.}
\label{alg:collab}
\begin{algorithmic}[1]
\Require Seed $\seed=(v_i,\tseed)$; fan-outs $M_1, M_2$; output size $K_{\mathrm{collab}}$.
\Ensure Fixed-length ID list $N_{\mathrm{collab}}$ of size $K_{\mathrm{collab}}$ with validity mask.
\State $E \gets M_1$ uniform samples from the causal neighbors of $v_i$
\State multiplicity $m \gets \varnothing$
\For{each entity $r \in E$}
  \For{each of $M_2$ uniform samples $p$ from the causal neighbors of $r$}
    \If{$p \neq v_i$ \textbf{ and } $p \notin E$ \textbf{ and } $p$ is not a $1$-hop neighbor of $v_i$}
      \State $m[p] \gets m[p] + 1$
    \EndIf
  \EndFor
\EndFor
\State $N_{\mathrm{collab}} \gets$ top $K_{\mathrm{collab}}$ peers by $m$; pad and mask as needed
\State \Return $N_{\mathrm{collab}}$
\end{algorithmic}
\end{algorithm}

\subsubsection{CRW Sampler: Walk Kernel, PPR Equivalence, Connectivity, and Edge Cases}
\label{app:ctrwr}
This appendix expands the CRW local subgraph sampler of \S\ref{sec:ctrwr} (Algorithm~\ref{alg:rwr}): the walk kernel, its convergence to causal PPR, the connectivity of the induced subgraph, and the fixed-length padding used when a seed's causal neighborhood is sparse.

\paragraph{Walk Kernel.}
Fix a seed $\seed=(v_i, \tseed)$ and a restart probability $p \in (0,1)$. To mitigate the fan-out from high-degree hubs, we restrict a given node $u$ to a subset of up to $M$ causal neighbors. While this truncation can be performed randomly, we default to a recency-based approach to prioritize temporally relevant interactions. Specifically, let $\mathcal{N}(u; \tseed)$ be the set of all temporal neighbors of node $u$ prior to the seed time:
$$ \mathcal{N}(u; \tseed) = \{ v \in V \mid (u,v) \in E,\, \tau(u,v) \le \tseed \} $$
We then define the sampled neighborhood $\mathcal{N}_M(u; \seed) \subseteq \mathcal{N}(u; \tseed)$ as the subset containing the $M$ nodes with the largest $\tau(u,v)$ values (i.e., the most recent interactions). If $|\mathcal{N}(u; \tseed)| < M$, all valid neighbors are kept.
At each step, a walk currently at node $u$ will restart to the seed with probability $p$, or otherwise transition uniformly to one of its $M$ valid causal neighbors:

\begin{equation}
P(u \to v \mid s) = 
\begin{cases}
\mathbb{I}[v = v_i], & \text{if } \mathcal{N}_M(u; s) = \varnothing, \\[4pt]
p \, \mathbb{I}[v = v_i] + (1-p) \, \frac{\mathbb{I}[v \in \mathcal{N}_M(u; s)]}{|\mathcal{N}_M(u; s)|}, & \text{otherwise.}
\end{cases}
\label{eq:ctrwr-kernel}
\end{equation}

If a dead end is reached where $\Nval_M(u;\seed)=\varnothing$, the walk automatically teleports back to $v_i$.
We execute $W$ independent walks of length $L$ from $v_i$, recording the total visit count $\mathrm{visit}[v]$ and the minimum discovery depth $\mathrm{depth}[v]$ for every visited node.

\textbf{Equivalence to Personalized PageRank.}
The Monte Carlo estimate of the visit distribution is governed by three parameters: the number of walks $W$, the maximum walk length $L$, and the branching truncation limit $M$. Normalizing the visit frequencies over these $W$ walks yields an empirical distribution $\hat{\pi}_{s}(v)$ that asymptotically converges to the exact Personalized PageRank (PPR) vector $\pi_{s}(v)$ of the seed $v_{i}$ when all parameters approach infinity:

\begin{equation}
    \hat{\pi}_{s}(v) = \frac{\text{visit}[v]}{\sum_{u} \text{visit}[u]} \xrightarrow{W,L,M\to\infty} \pi_{s}(v)
\end{equation}

In this limit, $L\to\infty$ eliminates the finite-horizon bias, $W\to\infty$ suppresses the Monte Carlo variance, and $M\to\infty$ recovers the exact transition probabilities of a standard random walk with restart. In practice, however, keeping $M$ finite serves as a crucial regularizer: it bounds per-step complexity independently of hub degree, and our default recency-based truncation injects a valuable temporal prior for time-dependent tasks.

\paragraph{Token Selection and Hop Labels.}
The seed fills slot $0$ and the remaining $K-1$ tokens are the non-seed nodes of highest visit count---their truncated-PPR rank.
If the walks yield fewer than $K-1$ unique nodes, each is kept once and the rest are resampled in proportion to visit count; a seed with no causal history instead draws uniformly from $V$ into a dedicated fallback hop bin.
Every non-fallback token is labeled $\min(\hbfs,\hwalk)$, the tighter of its BFS distance within the induced subgraph and its shallowest walk depth $\hwalk$.


\paragraph{Subgraph Connectivity.}
A natural concern with PageRank-based sampling on scale-free relational graphs is that high-degree hubs might act as probability sinks, dominating the token budget and inducing a topologically sparse or disconnected subgraph.
Three properties of the walk kernel~\eqref{eq:ctrwr-kernel} rule this out.
First, the restart probability $p$ imposes an exponential decay on the visit distribution with respect to geodesic distance from the seed, so probability mass cannot accumulate at a multi-hop hub without proportionately weighting the one-hop nodes that bridge the seed to it.
Second, the $M$-truncation bounds the branching factor at every step: capping the effective degree traversed during a walk curbs the combinatorial path explosion that otherwise skews the stationary distribution toward global hubs.
Third, whereas uniform BFS sampling at a high-degree node draws an essentially random subset of neighbors---typically inducing a disconnected, star-like subgraph with near-zero clustering---CRW aggregates visit frequencies across many short walks, a trace-based signal that intrinsically favors nodes sharing dense structural corridors with the seed.
The top-ranked $K-1$ tokens therefore form a cohesive subgraph with substantially higher internal edge density, preserving the structural integrity that the Local Module's self-attention (\S\ref{sec:local}) relies on.

\paragraph{Handling Sparse Neighborhoods and Edge Cases.}
Because all three branches (struct, temp and collab) require fixed-length sequences, the samplers must gracefully handle structural sparsity.
If a sampler extracts $n$ unique valid nodes where $0 < n < K-1$, it first retains all $n$ nodes and then fills the remaining $K-1-n$ sequence slots via resampling (using visit-count weights for CRW, and uniform weights for the temporal and collaborative branches).
In the extreme edge case where a seed possesses absolutely no valid causal history ($n=0$), the sequence is padded using nodes drawn uniformly from the entire graph $V$.
To strictly prevent artificial topological leakage, these uniform fallback tokens are explicitly assigned to a dedicated fallback hop-embedding bin.

\subsubsection{Collaborative Sampler: Two-Hop Probe as a Jaccard-Numerator Estimator}
\label{app:collab}
This appendix formalizes the claim of \S\ref{sec:collab-sampler} that the collaborative sampler's per-peer hit count (Algorithm~\ref{alg:collab}) is a Monte Carlo estimate of the Jaccard-overlap numerator between the seed's and a candidate peer's causal neighborhoods.

\paragraph{Causal neighborhoods and target quantity.}
For any node $u$, its causal neighbor set at the seed time and the corresponding degree are
$$ \Nval(u;\seed) = \bigl\{\, v : (u,v)\in E,\ \tau(u,v)\le\tseed \,\bigr\}, \qquad d_u = \bigl|\Nval(u;\seed)\bigr|. $$
Writing $\mathcal{S}(v_i,p)=\Nval(v_i;\seed)\cap\Nval(p;\seed)$ for the shared-intermediary set of the seed $v_i$ and a candidate peer $p$, the collaborative affinity we target is the numerator of their neighborhood Jaccard coefficient:
\begin{equation}
J(v_i,p) = \frac{\bigl|\mathcal{S}(v_i,p)\bigr|}{\bigl|\Nval(v_i;\seed)\cup\Nval(p;\seed)\bigr|},
\qquad
\mathrm{num}(v_i,p) = \bigl|\mathcal{S}(v_i,p)\bigr|.
\label{eq:collab-jaccard}
\end{equation}

\paragraph{Two-hop probe and hit count.}
Hop~1 draws a set $E$ of $M_1$ intermediaries uniformly without replacement from $\Nval(v_i;\seed)$, and for each $r\in E$, Hop~2 draws $M_2$ peers uniformly without replacement from $\Nval(r;\seed)$.
After discarding the seed, the Hop-1 set, and the seed's one-hop neighbors, every surviving peer accumulates the hit count
\begin{equation}
m[p] = \sum_{r\in E}\mathbb{I}\bigl[\,p\in \mathrm{Hop2}(r)\,\bigr],
\qquad p\notin \{v_i\}\cup\Nval(v_i;\seed),
\label{eq:collab-hits}
\end{equation}
which is nonzero only for shared intermediaries $r\in\mathcal{S}(v_i,p)$, since $p\in\Nval(r;\seed)\Leftrightarrow r\in\Nval(p;\seed)$.

\paragraph{Expected hit count.}
Under simple random sampling without replacement the marginal inclusion probabilities are $P(r\in E)=M_1/d_{v_i}$ and $P\bigl(p\in\mathrm{Hop2}(r)\mid r\in E\bigr)=M_2/d_r$ for $p\in\Nval(r;\seed)$, so for budgets $M_1\le d_{v_i}$ and $M_2\le d_r$
\begin{equation}
\mathbb{E}\bigl[m[p]\bigr]
= \sum_{r\in\mathcal{S}(v_i,p)} \frac{M_1}{d_{v_i}}\cdot\frac{M_2}{d_r}
= \frac{M_1 M_2}{d_{v_i}}\sum_{r\in\mathcal{S}(v_i,p)}\frac{1}{d_r}.
\label{eq:collab-expectation}
\end{equation}

\paragraph{Proportionality to the Jaccard numerator.}
Rescaling \eqref{eq:collab-expectation} by the peer-independent seed constant $d_{v_i}/(M_1 M_2)$ yields an unbiased estimator of a degree-discounted common-neighbor count:
\begin{equation}
\hat{c}(p) = \frac{d_{v_i}}{M_1 M_2}\,m[p],
\qquad
\mathbb{E}\bigl[\hat{c}(p)\bigr] = \sum_{r\in\mathcal{S}(v_i,p)}\frac{1}{d_r}.
\label{eq:collab-ra}
\end{equation}
When shared intermediaries have comparable degree $d_r\approx\bar d$, this collapses to a scaled Jaccard numerator,
\begin{equation}
\mathbb{E}\bigl[m[p]\bigr] \approx \frac{M_1 M_2}{d_{v_i}\,\bar d}\,\bigl|\mathcal{S}(v_i,p)\bigr| \;\propto\; \mathrm{num}(v_i,p).
\label{eq:collab-proportional}
\end{equation}
Since $d_{v_i}$ is fixed across all peers of a given seed, ranking candidates by $m[p]$ and keeping the top $K_{\mathrm{collab}}$ (Algorithm~\ref{alg:collab}) selects the peers of largest expected neighborhood overlap, with the $1/d_r$ weighting additionally down-weighting popular high-degree intermediaries.

\subsubsection{Token Encoder Details}
\label{app:token-encoders}
This appendix gives the exact per-encoder formulas for the 5-tuple token construction summarized in \S\ref{sec:tokenization}.
Each of the $K$ sampled tokens $v_j$ is processed by the following specialized encoders before integration:

\begin{enumerate}
    \item \textbf{Node Feature Encoder:} Raw columnar attributes $x_{v_j}$ are encoded into a $d$-dimensional embedding~\citep{gorishniy2021revisiting,hollmann2023tabpfn,kim2024carte} using a PyTorch Frame~\citep{hu2024pytorchframe} multimodal encoder, which processes and aggregates numerical, categorical, and text modalities:
    \begin{equation}
    h_{\mathrm{feat}}(v_j) = \mathrm{MultiModalEncoder}(x_{v_j}) \in \mathbb{R}^d
    \label{eq:hfeat}
    \end{equation}

    \item \textbf{Node Type Encoder:} Converts the table-specific entity type $\phi(v_j)$ into a learned representation to capture heterogeneous schema semantics:
    \begin{equation}
    h_{\mathrm{type}}(v_j) = W_{\mathrm{type}} \cdot \mathrm{onehot}(\phi(v_j)) \in \mathbb{R}^d
    \end{equation}
    where $W_{\mathrm{type}} \in \mathbb{R}^{d \times |\mathcal{T}|}$ is a learnable weight matrix.

    \item \textbf{Hop Encoder:} Captures the structural proximity $p(v_i, v_j)$ between the seed node $v_i$ and its neighbor $v_j$:
    \begin{equation}
    h_{\mathrm{hop}}(v_i, v_j) = W_{\mathrm{hop}} \cdot \mathrm{onehot}(p(v_i, v_j)) \in \mathbb{R}^d
    \end{equation}
    Rather than a single shortest-path distance, the label is the tighter of two causal upper bounds---the BFS distance within the induced $K$-token subgraph and the CRW walk depth:
    \begin{equation}
    p(v_i, v_j) = \min\bigl(\hbfs(v_i, v_j),\ \hwalk(v_j)\bigr),
    \end{equation}
    with a dedicated bin reserved for uniformly-drawn fallback tokens.
    \item \textbf{Time Encoder:} Linearly transforms the relative temporal difference to ensure temporal alignment:
    \begin{equation}
    h_{\mathrm{time}}(v_i, v_j) = W_{\mathrm{time}} \cdot (\tau(v_j) - \tau(v_i)) \in \mathbb{R}^d
    \end{equation}
    where $W_{\mathrm{time}} \in \mathbb{R}^d$ is a learnable parameter.

    \item \textbf{Subgraph Positional Encoder (PE):} Captures local graph topology (e.g., cycles, parent-child relationships) by applying a lightweight GIN to the sampled local subgraph's adjacency matrix $A_{\mathrm{local}}$.
To break structural symmetries while preserving permutation equivariance, we inject stochastically resampled random node features $Z_{\mathrm{random}}$ at every training step following the convention in~\citep{dwivedi2026relgt}:
    \begin{equation}
    h_{\mathrm{pe}}(v_j) = \mathrm{GIN}(A_{\mathrm{local}}, Z_{\mathrm{random}})_j \in \mathbb{R}^d
    \label{eq:GIN}
    \end{equation}
\end{enumerate}

The final token representation is formed by concatenating the five encoded elements and mixing them via a learned projection matrix $W_{\mathrm{mix}} \in \mathbb{R}^{d \times 5d}$:
\begin{equation}
h_{\mathrm{token}}(v_j) = W_{\mathrm{mix}} \cdot [h_{\mathrm{feat}}(v_j) \parallel h_{\mathrm{type}}(v_j) \parallel h_{\mathrm{hop}}(v_i, v_j) \parallel h_{\mathrm{time}}(v_i, v_j) \parallel h_{\mathrm{pe}}(v_j)]
\label{eq:htoken}
\end{equation}

\subsection{Complexity Analysis}
\label{app:complexity}

\subsubsection{Subgraph Sampler Time Complexity}
\label{sec:complexity-samplers}
Table~\ref{tab:complexity} reports the per-seed cost of the three causal samplers of \S\ref{sec:sampling}.
The CRW sampler (\S\ref{sec:ctrwr}) runs $W$ independent restart walks of length $L$ from the seed, giving a per-seed cost of $O(WL\log D_{\max})$, where $D_{\max}$ is the maximum node degree and the $\log D_{\max}$ factor is the per-step causal-neighbor lookup.
Under the default parameters ($W=100$, $L=10$, $p=0.15$, $M=50$, $K=300$) this avoids linear dependence on hub degree, since the recency-based $M$-truncation caps the per-step branching regardless of hub degree.
The collaborative sampler (\S\ref{sec:collab-sampler}) draws $M_1$ first-hop intermediaries and $M_2$ second-hop peers from each, for a per-seed cost of $O(M_1 M_2\log D_{\max})$.
The temporal sampler (\S\ref{sec:temporal-sampler}) scans a seed-independent, global time-sorted event stream with type-balanced budgets: for each of the $T=|\mathcal{T}|$ node types it binary-searches the stream to locate the causal window ($O(\log|E|)$, with $|E|$ the number of REG edges), then collects the $B=K_{\mathrm{temp}}/T$ most-recent nodes per type, giving a per-seed cost of $O(T\log|E| + TB)$.
Because the candidate pool is global, this cost is independent of the seed's degree.
Crucially, because all extracted subgraphs and global neighbor sequences are pre-computed and cached offline, this cost is heavily amortized; it is paid exactly once prior to training, adding zero traversal overhead to the active training loop.\begin{table}[t]
\centering
\small
\caption{Per-seed time complexity of the three causal subgraph samplers (\S\ref{sec:sampling}). $D_{\max}$: maximum node degree; $T=|\mathcal{T}|$: number of node types; $B=K_{\mathrm{temp}}/T$: per-type temporal budget; $|E|$: number of REG edges.}
\label{tab:complexity}
\begin{tabular}{ll}
\toprule
Sampler & Per-seed cost \\
\midrule
Local (\S\ref{sec:ctrwr}) & $O(WL \log D_{\max})$ \\
Collab.\ (\S\ref{sec:collab-sampler}) & $O(M_1 M_2 \log D_{\max})$ \\
Temporal (\S\ref{sec:temporal-sampler}) & $O(T \log |E| + T B)$ \\
\bottomrule
\end{tabular}
\end{table}

\subsubsection{Attention Module Time Complexity}
\label{sec:complexity-attention}
This appendix collects the per-seed costs of \textsc{Quartet}'s attention modules: the local self-attention module of \S\ref{sec:local} and the four global branches of \S\ref{sec:global}.

\paragraph{Local attention module.}
The local module applies $L_{\mathrm{local}}$ layers of full self-attention over the fixed $K$-token subgraph, costing $O(L_{\mathrm{local}} K^2 d)$ per seed.
Because the CRW sampler holds $K$ constant regardless of hub degree (\S\ref{sec:ctrwr}), this cost is fixed across seeds and datasets rather than scaling with local density.

\paragraph{Feature and topological codebook branches.}
Both codebook reads (\S\ref{sec:codebooks}) are inexpensive because the archetypes are batch-shared; let $R = R_{\mathrm{struct}} = R_{\mathrm{feat}}$ denote the shared codebook size.
The key and value projections $W_K E_b$ and $W_V E_b$ cost $O(Rd^2)$ and are computed once per batch and reused across all seeds, rather than recomputed for every seed as in a naive codebook lookup; each seed then adds only an $O(Rd)$ query--archetype attention.
The shared projections likewise cap the key/value activation memory at $O(Rd)$ instead of letting it grow linearly with the number of seeds---saving roughly $0.5$ GB per codebook when the batch size, archetype count $R$, and width $d$ are all $512$.

\paragraph{Temporal and collaborative cross-attention branches.}
The temporal and collaborative branches (\S\ref{sec:xatt}) replace the $O(K^2 d)$ cost of full self-attention over their length-$K$ context $\mathbf{C}_b$ ($b \in \{\mathrm{temp}, \mathrm{collab}\}$) with an asymmetric Perceiver bottleneck: a fixed budget of $L_{\mathrm{perc}} \ll K$ latents cross-attends over the context, reducing each branch's cost to $O(L_{\mathrm{perc}} K d)$---linear in the context length.
With $L_{\mathrm{perc}}=32$ and $K=300$ this holds each branch's compute and its fusion-input width constant regardless of how large the underlying temporal or collaborative neighborhood is.

\subsubsection{Empirical Runtime, Memory, and Storage Comparison}
\label{sec:complexity-empirical}
To complement the asymptotic analysis above with a direct empirical measurement, we profile \textsc{Quartet} against its RelGT base on the five larger classification tasks of \S\ref{sec:impl} (Table~\ref{tab:hp-per-task}), holding batch size, sampler token budgets, and local-layer depth fixed across both models and running each on a single A100 GPU per model to keep the comparison unconfounded by DDP world size. Table~\ref{tab:efficiency} reports, for every task, the ratio of \textsc{Quartet}'s cost to RelGT's along five axes: per-epoch train+val wall-clock, inference throughput, one-time offline precomputation, peak GPU memory, on-disk cache size, and parameter count.

\paragraph{Parameters and memory scale with the quad-branch capacity, as expected.}
Parameter count sits in a tight $2.17${--}$2.30\times$ band across every task, consistent with it being a fixed property of the four global branches' added capacity (\S\ref{sec:global}) rather than a task-dependent quantity. Peak GPU memory follows the same architectural source---the codebook and Perceiver branches' batch-shared buffers (\S\ref{sec:complexity-attention})---and is higher on every task, averaging $2.58\times$ RelGT's. On-disk cache size is the largest relative cost, averaging $7.76\times$, because it accumulates the additional per-seed temporal and collaborative neighbor sequences that RelGT's local-only sampler does not cache.

\paragraph{Runtime is favorable on average, but we do not attribute it primarily to the samplers' asymptotic advantage.}
Per-epoch time is $1.39\times$ RelGT's on average (as low as $0.84\times$ on \texttt{rel-amazon/user-churn}), and inference throughput favors \textsc{Quartet} on three of the five tasks, most on the tasks with the largest test sets. Offline precomputation is faster for \textsc{Quartet} on four of five tasks, by $1.3${--}$9.7\times$. While \S\ref{sec:complexity-samplers} shows CRW's per-seed cost is bounded independently of hub degree, we traced this specific empirical gap to our profiling harness and found it is driven substantially by implementation differences between the two reproduction pipelines---in particular, an unguarded multi-process precompute step in the RelGT baseline that re-derives its adjacency structure from scratch on every worker and every data chunk, rather than once per split. These differences are orthogonal to the sampling algorithm and would likely narrow this specific gap if fixed on the baseline side; we therefore report the measured numbers without treating them as direct validation of the asymptotic argument in \S\ref{sec:complexity-samplers}. The inference-throughput advantage has a more direct architectural explanation: RelGT's EMA-K-Means global module recomputes each centroid's current occupancy via \texttt{unique()} on every forward pass, an operation whose output size is data-dependent and therefore forces a GPU--host synchronization on every batch, at both train and test time; \textsc{Quartet}'s codebook and Perceiver branches (\S\ref{sec:codebooks}, \S\ref{sec:xatt}) are static-shape tensor operations throughout and incur no equivalent stall. This cost is a small fraction of a backward-dominated training step but a much larger fraction of a short, forward-only inference step, which is consistent with the throughput advantage being largest on the tasks with the most test batches and narrowing on the smallest task in our set.\begin{table}[t]
\centering
\small
\caption{Empirical runtime, memory, and storage of \textsc{Quartet} relative to RelGT (ratio \textsc{Quartet}/RelGT), measured on the five larger classification tasks of \S\ref{sec:impl} (Table~\ref{tab:hp-per-task}), each run on a single A100 GPU per model for a matched, unconfounded comparison. Epoch time is mean per-epoch train+val wall-clock; throughput is test-set samples/sec at inference; precompute is the one-time offline sampling/caching pass; cache is total on-disk size of the pre-cached artefacts. For Epoch time, Precompute, Peak mem, Cache, and Params, $>1\times$ means \textsc{Quartet} costs more; for Throughput, $>1\times$ means \textsc{Quartet} is faster.}
\label{tab:efficiency}
\begin{tabular}{lcccccc}
\toprule
Task & Epoch time & Throughput & Precompute & Peak mem & Cache & Params \\
\midrule
rel-hm/user-churn         & $1.01\times$ & $1.74\times$ & $0.10\times$ & $1.70\times$ & $3.60\times$ & $2.23\times$ \\
rel-stack/user-engagement & $1.83\times$ & $1.09\times$ & $1.04\times$ & $3.07\times$ & $8.02\times$ & $2.17\times$ \\
rel-amazon/user-churn     & $0.84\times$ & $2.09\times$ & $0.31\times$ & $2.52\times$ & $8.72\times$ & $2.30\times$ \\
rel-amazon/item-churn     & $1.51\times$ & $1.29\times$ & $0.36\times$ & $2.54\times$ & $8.75\times$ & $2.30\times$ \\
rel-stack/user-badge      & $1.77\times$ & $0.96\times$ & $0.75\times$ & $3.06\times$ & $9.69\times$ & $2.17\times$ \\
\midrule
Mean                      & $1.39\times$ & $1.43\times$ & $0.51\times$ & $2.58\times$ & $7.76\times$ & $2.23\times$ \\
\bottomrule
\end{tabular}
\end{table}

\paragraph{The added cost is a dial, not a fixed tax.}
GPU memory and cache storage are the costs we attribute with the most confidence to our architecture, and both scale specifically with the four global branches rather than with the local module or samplers. Crucially, \textsc{Quartet} is not a single fixed-cost model but a family of configurations---the full quad-branch model, or any subset of its branches---and the leave-one-out ablation of \S\ref{sec:results-ablation} already quantifies, per task, which branches are worth their share of this cost: no single branch is globally redundant, as the dominant branch shifts across tasks (\S\ref{sec:results-ablation}). A practitioner can therefore pay only for the branch(es) that carry signal for their specific downstream task and recover most of RelGT's memory and storage footprint elsewhere, rather than weighing the full-model cost against the full-model gain.

\subsection{Experiment Details}
\label{app:experiment-details}

\subsubsection{Full Implementation and Training Details}
\label{sec:impl}
All models are trained end-to-end with Adam at a constant learning rate of $1{\times}10^{-4}$, batch size 512, and gradient clipping at $1.0$; we apply no learning-rate warmup or decay schedule.
The training budget is set per task: 100 epochs for the seven smaller tasks and 10 epochs for the five larger tasks.
Local-transformer width follows the RelGT defaults~\citep{dwivedi2026relgt} ($c = 512$ channels, $h = 4$ heads).
The number of local layers $L_{\mathrm{local}}$, dropout (tied across the feed-forward and attention paths), and weight decay are tuned per task: on the seven smaller tasks we search a $3{\times}3{\times}3$ grid of $L_{\mathrm{local}} \in \{1, 4, 8\}$, dropout $\in \{0.3, 0.4, 0.5\}$, and weight decay $\in \{5{\times}10^{-5}, 5{\times}10^{-4}, 5{\times}10^{-3}\}$ (reproducing RelGT's $L{\times}$dropout grid and adding a weight-decay axis), while on the five larger tasks we fix $L_{\mathrm{local}} = 4$ and search a $3{\times}3$ grid of dropout $\in \{0.1, 0.3, 0.5\}$ and weight decay $\in \{1{\times}10^{-5}, 5{\times}10^{-5}, 5{\times}10^{-4}\}$.
The per-task best configurations are reported in Appendix~\ref{app:hp}.
These local hyperparameters were identified on the CRW-augmented RelGT base (\S\ref{sec:results-rwr}) and held fixed during the subsequent quad-branch capacity search.
CRW uses $W = 100$ walks, $L = 10$ steps, restart probability $p = 0.15$, recent-$M=50$, and token budget $K = 300$.

Quad-branch capacity---Perceiver latents $L_{\mathrm{perc}} \in \{16, 32, 64\}$ and codebook archetypes $R_{\mathrm{struct}} = R_{\mathrm{feat}} \in \{128, 256, 512\}$---is configured per task, with the shared self-attention depth held fixed at $L_{\mathrm{sa}} = 4$ and the branch outputs fused by concatenation (the gate-free \texttt{nogate\_concat} default of \S\ref{sec:fusion}); the full per-arm capacity results are tabulated in Appendix~\ref{app:hp-capacity}.
The collaborative sampler uses $(M_1, M_2, K_{\mathrm{collab}}) = (50, 50, 300)$, and the temporal branch draws its $K_{\mathrm{temp}} = 300$ tokens using the type-balanced sampler of \S\ref{sec:temporal-sampler} (Algorithm~\ref{alg:temporal}) with per-type budget $B = K_{\mathrm{temp}}/|\mathcal{T}|$ over the $\Delta t = 365$-day causal window; the candidate pool is global (seed-independent), but the per-type budget ensures no single high-frequency node type dominates the sample.
Both cross-attention branches thus operate at the same $K = 300$ token budget as the CRW local context.
The pre-cache pipeline writes the four artifacts per seed to a single HDF5 file atomically ($\mathtt{tmp}$-file + $\mathtt{os.rename}$) and is consumed at train time via an $\mathtt{mmap}$-backed loader that performs no on-line graph traversal.
Checkpoint selection is done by best validation ROC-AUC.

\subsubsection{Hyperparameter Tuning Strategy and Baseline Comparisons}
\label{app:hp-strategy}

To ensure a fair and computationally realistic comparison, we adopted a deliberate hyperparameter tuning strategy across all evaluated models.
We directly applied the published optimal hyperparameters for HGT and RelGT.
Re-tuning was deemed unnecessary because the classification datasets in \textsc{RelBench}~v2 are unchanged to the prior version, aside from a patch to rel-event user-ignore that resolved temporal leakage.
For \textsc{Quartet}, we avoided a computationally prohibitive joint grid search by employing a constrained, decoupled two-stage approach:

\paragraph{Stage~1 (Local Parameters).}
We first optimized the local module's parameters ($L_{\mathrm{local}}$, dropout, and weight decay).
Weight decay was included as an additional dimension specifically because the CRW sampler extracts denser, more highly connected subgraphs, which occasionally necessitated additional regularization to prevent overfitting.

\paragraph{Stage~2 (Global Parameters).}
Once the optimal local parameters were identified and frozen, we conducted a separate, limited search over the global branch capacities (Perceiver latents $L_{\mathrm{perc}}$ and codebook size $R$).

This decoupled strategy significantly restricted \textsc{Quartet}'s total search budget compared to an exhaustive combinatorial grid, ensuring a fair and equitable comparison against the author-optimized baselines from the literature.

\subsubsection{Per-Task Hyperparameters}
\label{app:hp}
Table~\ref{tab:hp-per-task} lists the best per-task configuration of the three tuned local hyperparameters---number of local layers $L_{\mathrm{local}}$, dropout, and weight decay---for all twelve classification tasks, selected over the grids described in \S\ref{sec:impl}.
All remaining local hyperparameters are shared across tasks (\S\ref{sec:impl}).
\begin{table}[t]
\centering
\small
\caption{
  Per-task best hyperparameters selected by the per-task tuning of \S\ref{sec:impl}:
  number of local layers $L_{\mathrm{local}}$, dropout, and weight decay.
  Dropout is applied identically to the feed-forward and attention paths.
  The seven smaller tasks search $L_{\mathrm{local}} \in \{1, 4, 8\}$; the five
  larger tasks fix $L_{\mathrm{local}} = 4$.
}
\label{tab:hp-per-task}
\begin{tabular}{llccc}
\toprule
Dataset & Task & $L_{\mathrm{local}}$ & Dropout & Weight decay \\
\midrule
\multicolumn{5}{l}{\emph{Smaller tasks} ($3{\times}3{\times}3$ grid)} \\
rel-avito  & user-clicks     & 1 & 0.4 & $5{\times}10^{-3}$ \\
           & user-visits     & 8 & 0.5 & $5{\times}10^{-5}$ \\
\midrule
rel-event  & user-ignore     & 1 & 0.3 & $5{\times}10^{-5}$ \\
           & user-repeat     & 4 & 0.5 & $5{\times}10^{-3}$ \\
\midrule
rel-f1     & driver-dnf      & 4 & 0.4 & $5{\times}10^{-5}$ \\
           & driver-top3     & 1 & 0.5 & $5{\times}10^{-3}$ \\
\midrule
rel-trial  & study-outcome   & 1 & 0.5 & $5{\times}10^{-4}$ \\
\midrule
\multicolumn{5}{l}{\emph{Larger tasks} ($3{\times}3$ grid, $L_{\mathrm{local}}$ fixed)} \\
rel-amazon & item-churn      & 4 & 0.1 & $5{\times}10^{-5}$ \\
           & user-churn      & 4 & 0.5 & $5{\times}10^{-5}$ \\
\midrule
rel-hm     & user-churn      & 4 & 0.3 & $1{\times}10^{-5}$ \\
\midrule
rel-stack  & user-badge      & 4 & 0.5 & $1{\times}10^{-5}$ \\
           & user-engagement & 4 & 0.1 & $1{\times}10^{-5}$ \\
\bottomrule
\end{tabular}
\end{table}

\subsubsection{Quad-Branch Capacity
Search}
\label{app:hp-capacity}
Table~\ref{tab:hp-capacity} reports validation and test ROC-AUC for every arm of the quad-branch capacity grid of \S\ref{sec:impl}---Perceiver latent budget $L_{\mathrm{perc}}\in\{16,32,64\}$ crossed with codebook size $R_{\mathrm{struct}}=R_{\mathrm{feat}}\in\{128,256,512\}$---on all twelve classification tasks, averaged over $n=4$ seeds.
For each task the bold entry marks the configuration reported in the \textsc{Quartet} column of Table~\ref{tab:baselines}, with validation scores listed alongside.
\begin{table}[t]
\centering
\small
\setlength{\tabcolsep}{3pt}
\caption{
  Quad-branch capacity search (companion to \S\ref{sec:impl}): validation and test
  ROC-AUC on the twelve \textsc{RelBench} classification tasks across the
  $L_{\mathrm{perc}}\times R$ capacity grid. Column groups index the Perceiver latent
  budget $L_{\mathrm{perc}}\in\{16,32,64\}$; sub-columns (128/256/512) index the codebook
  size $R=R_{\mathrm{struct}}=R_{\mathrm{feat}}$. Each arm is averaged over $n=4$ seeds.
  \textbf{Bold} marks the configuration reported in the \textsc{Quartet} column of
  Table~\ref{tab:baselines}; validation scores are listed alongside for completeness.
}
\label{tab:hp-capacity}
\resizebox{\textwidth}{!}{%
\begin{tabular}{lllccccccccc}
\toprule
\multirow{2}{*}{Dataset} & \multirow{2}{*}{Task} & \multirow{2}{*}{AUC\,$\uparrow$}
 & \multicolumn{3}{c}{$L_{\mathrm{perc}}=16$}
 & \multicolumn{3}{c}{$L_{\mathrm{perc}}=32$}
 & \multicolumn{3}{c}{$L_{\mathrm{perc}}=64$} \\
\cmidrule(lr){4-6}\cmidrule(lr){7-9}\cmidrule(lr){10-12}
 & & & 128 & 256 & 512 & 128 & 256 & 512 & 128 & 256 & 512 \\
\midrule
rel-amazon & item-churn & Test & 0.8252 & 0.8246 & 0.8259 & 0.8244 & 0.8255 & \textbf{0.8262} & 0.8256 & 0.8249 & 0.8248 \\
           &            & Val  & 0.8217 & 0.8215 & 0.8221 & 0.8211 & 0.8221 & 0.8227 & 0.8219 & 0.8214 & 0.8213 \\
           & user-churn & Test & 0.6995 & 0.6983 & 0.6976 & 0.6969 & 0.6988 & \textbf{0.7001} & 0.6962 & 0.6977 & 0.6983 \\
           &            & Val  & 0.7005 & 0.6999 & 0.6987 & 0.6985 & 0.6998 & 0.7003 & 0.6980 & 0.6989 & 0.6992 \\
\midrule
rel-avito & user-clicks & Test & 0.5624 & 0.6268 & 0.6367 & \textbf{0.6481} & 0.6298 & 0.6072 & 0.6209 & 0.6089 & 0.6365 \\
          &             & Val  & 0.5591 & 0.6148 & 0.6055 & 0.6087 & 0.6110 & 0.5988 & 0.6115 & 0.5935 & 0.6065 \\
          & user-visits & Test & 0.6291 & 0.6305 & 0.6245 & 0.6378 & 0.6307 & 0.6402 & 0.6422 & 0.6391 & \textbf{0.6508} \\
          &             & Val  & 0.6738 & 0.6792 & 0.6769 & 0.6818 & 0.6783 & 0.6851 & 0.6821 & 0.6834 & 0.6892 \\
\midrule
rel-event & user-ignore & Test & 0.7296 & \textbf{0.7392} & 0.7345 & 0.7330 & 0.7245 & 0.7339 & 0.7337 & 0.7270 & 0.7282 \\
          &             & Val  & 0.8092 & 0.8176 & 0.8061 & 0.8086 & 0.8076 & 0.8119 & 0.8143 & 0.8066 & 0.8111 \\
          & user-repeat & Test & \textbf{0.7563} & 0.7493 & 0.7496 & 0.7401 & 0.7438 & 0.7387 & 0.7311 & 0.7492 & 0.7356 \\
          &             & Val  & 0.7292 & 0.7323 & 0.7228 & 0.7319 & 0.7337 & 0.7399 & 0.7327 & 0.7267 & 0.7323 \\
\midrule
rel-f1 & driver-dnf  & Test & 0.6798 & 0.6717 & 0.6779 & 0.6753 & 0.6773 & 0.6941 & \textbf{0.7044} & 0.6711 & 0.6809 \\
       &             & Val  & 0.8104 & 0.8091 & 0.8040 & 0.8084 & 0.8060 & 0.8088 & 0.8040 & 0.8067 & 0.8074 \\
       & driver-top3 & Test & 0.7855 & 0.8042 & 0.8427 & 0.8391 & 0.8348 & 0.8265 & 0.8188 & \textbf{0.8517} & 0.8180 \\
       &             & Val  & 0.8733 & 0.8756 & 0.8835 & 0.8791 & 0.8697 & 0.8704 & 0.8735 & 0.8741 & 0.8715 \\
\midrule
rel-hm & user-churn & Test & 0.6967 & 0.6949 & 0.6960 & 0.6956 & 0.6957 & 0.6960 & 0.6947 & \textbf{0.6976} & 0.6957 \\
       &            & Val  & 0.7029 & 0.7011 & 0.7023 & 0.7022 & 0.7023 & 0.7017 & 0.7015 & 0.7033 & 0.7018 \\
\midrule
rel-stack & user-badge & Test & 0.8287 & 0.8093 & 0.8201 & 0.8159 & 0.8213 & 0.8300 & 0.8199 & 0.8344 & \textbf{0.8528} \\
          &            & Val  & 0.8342 & 0.8097 & 0.8234 & 0.8184 & 0.8282 & 0.8340 & 0.8265 & 0.8413 & 0.8580 \\
          & user-engagement & Test & 0.9053 & 0.9058 & 0.9050 & 0.9053 & 0.9054 & 0.9050 & 0.9057 & \textbf{0.9063} & 0.9057 \\
          &                 & Val  & 0.9019 & 0.9020 & 0.9023 & 0.9020 & 0.9021 & 0.9017 & 0.9025 & 0.9029 & 0.9028 \\
\midrule
rel-trial & study-outcome & Test & 0.6963 & 0.6995 & \textbf{0.7059} & 0.6986 & 0.6959 & 0.6962 & 0.6783 & 0.7015 & 0.6975 \\
          &               & Val  & 0.6756 & 0.6686 & 0.6746 & 0.6755 & 0.6732 & 0.6716 & 0.6687 & 0.6772 & 0.6738 \\
\bottomrule
\end{tabular}
}
\end{table}

\subsection{Additional Results}
\label{app:additional-results}

\subsubsection{Subgraph Connectivity Analysis: CRW extracts denser, more connected subgraphs compared to BFS2}
\label{app:crw-bfs-structure}

In this section, we provide further evidence to support the claim that Causal Random Walk (CRW) sampler yields denser and tightly connected subgraphs compared to the BFS2 baseline in RelGT.
We start with (I) a visual comparison of the two samplers, then (II) break down the graph-level statistics for the induced $K$-token subgraphs ($K=300$, retaining causal edges) across all 12 classification tasks and training seeds.

\begin{figure*}[t]
\centering
\includegraphics[width=\textwidth]{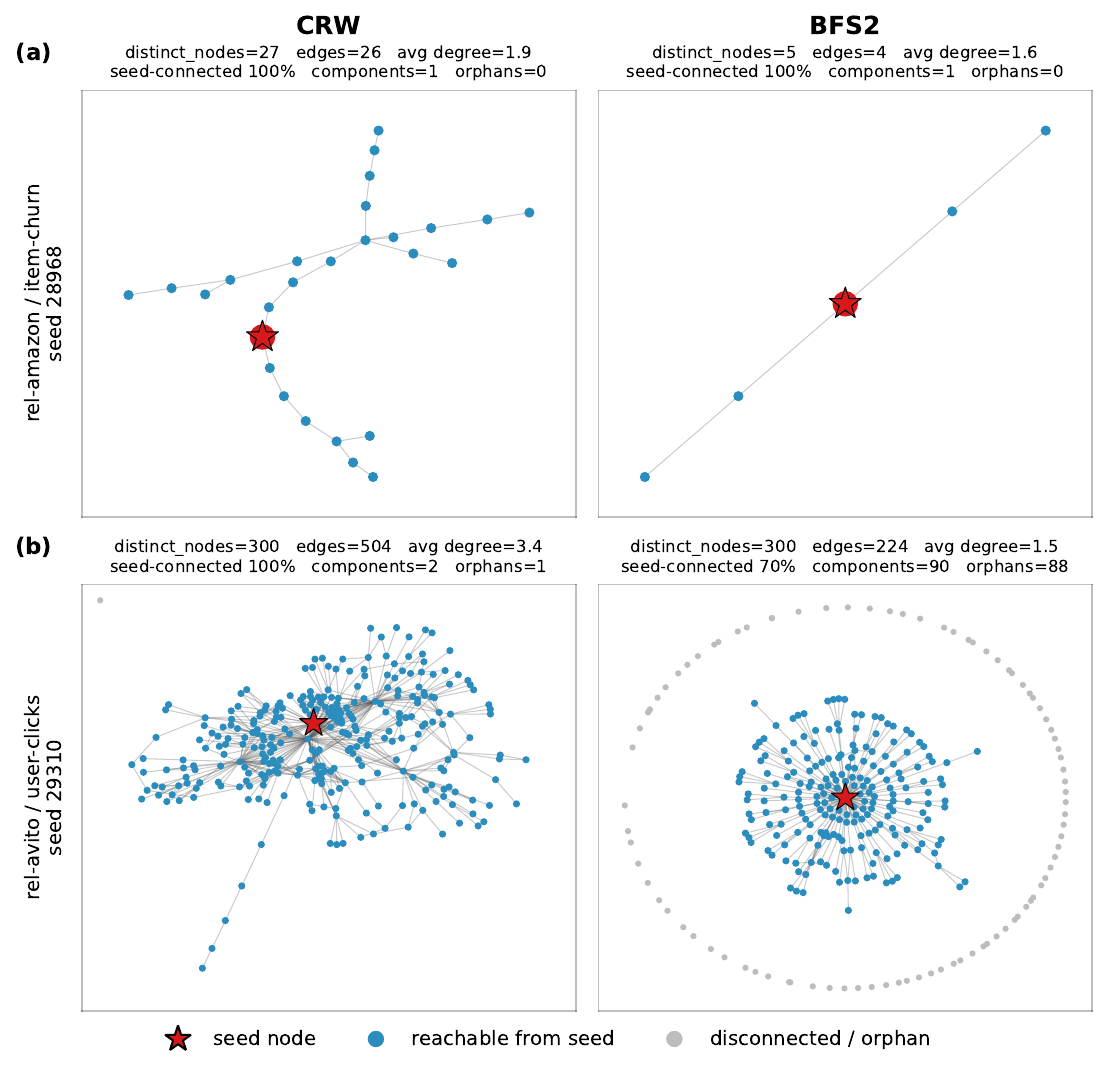}
\caption{Subgraph visualizations comparing CRW and BFS2 across two representative seeds:
\textbf{(a)}~A sparse-neighborhood seed.
\textbf{(b)}~A dense-neighborhood seed.
Red stars mark the seed node; blue nodes are reachable from the seed within the subgraph; grey nodes are disconnected orphans.
Note that the disconnected grey nodes in~(b) for BFS2 receive degenerate GIN positional encodings (Eq.~21), as no message-passing paths exist to differentiate them structurally.}
\label{fig:crw-bfs-examples}
\end{figure*}

\paragraph{(I) Visual intuition.}
Figure~\ref{fig:crw-bfs-examples} contrasts CRW and BFS2 subgraphs on two representative seeds that illustrate complementary regimes.
Panel~(a) shows a \emph{sparse-neighborhood} seed, with low causal degree.
Because the seed possesses a low causal degree, BFS2 exhausts its two-hop frontier after isolating only 5 distinct nodes. This leaves 295 of the 300 allocated budget slots wasted on duplicates and padding.
Conversely, CRW’s restart-walk mechanism successfully explores beyond this shallow frontier to extract 27 distinct nodes, forming a single connected component with an average degree of $\bar{d}=1.9$ and $100\%$ seed connectivity.
Panel~(b) shows a \emph{dense-neighborhood} seed.
While both samplers successfully draw 300 distinct nodes, the resulting structural topologies are drastically different. CRW builds a dense, seed-anchored cluster comprising only 2 components ($\bar{d}=3.4$, $\approx 100\%$ seed-connected, with only one disconnected node).
In contrast, BFS2 fragments the neighborhood into 90 disconnected components and leaves 88 nodes orphaned, resulting in a mere $70\%$ seed connectivity.
This severe fragmentation occurs because the BFS2 sampler selects two-hop neighbors but omits the crucial one-hop bridging nodes needed to connect them to the seed.

The disconnected grey nodes in the BFS2 panel in Figure \ref{fig:crw-bfs-examples}(b) are unreachable from the seed, representing a wasted token budget that cannot directly contribute to local message passing.
This structural fragmentation creates two distinct downstream bottlenecks.
First, it directly degrades the GIN-based positional encoder (Eq.\ref{eq:GIN}; \S\ref{sec:tokenization}).
Because GIN propagates information strictly along the edges of the subgraph adjacency matrix $A_{\mathrm{local}}$, these orphan nodes receive no neighborhood signals and collapse into identical positional encodings, eliminating the structural disambiguation the encoder is meant to provide.
Second, this collapse severely handicaps the Local Attention Module in Figure \ref{fig:arch}.
Even though the module applies full all-pair self-attention and can theoretically route information between any two tokens, the degraded positional encodings strip away crucial structural priors.
To compensate for the topological context lost during BFS2 sampling, the Transformer is forced to rely on additional self-attention layers just to infer the basic structural relationships between these disconnected nodes.

\begin{figure*}[t]\centering\includegraphics[width=\textwidth]{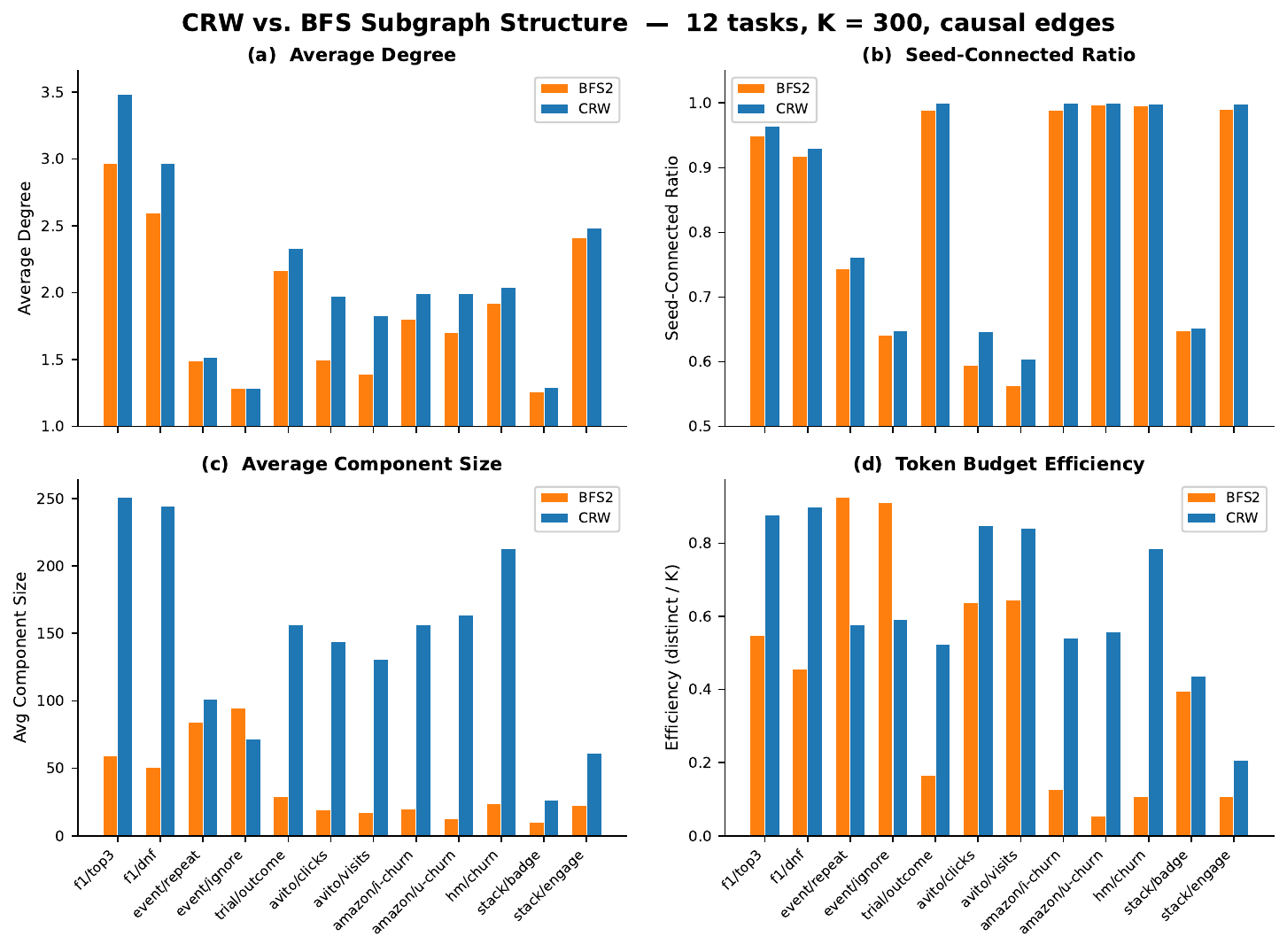}\caption{CRW vs.\ BFS2 subgraph structure across all 12 \textsc{RelBench} classification tasks ($K{=}300$, causal edges).Error bars show $\pm1$ standard deviation across training seeds.\textbf{(a)}~Average degree: CRW wins on all 12 tasks.\textbf{(b)}~Seed-connected ratio: CRW wins on all 12 tasks.\textbf{(c)}~Average component size: CRW produces larger components on all 12 tasks, reflecting greater structural cohesion.\textbf{(d)}~Token-budget efficiency: CRW is more efficient on 10 of 12 tasks; the two rel-event exceptions are explained by extreme seed fan-out.}\label{fig:crw-bfs-panel}\end{figure*}

\paragraph{(II) Aggregate Graph Metrics.}
Figure~\ref{fig:crw-bfs-panel} visualizes four key connectivity metrics across all twelve tasks to illustrate the structural differences between the samplers.
Panel~(a) demonstrates that CRW consistently achieves a higher average degree, meaning each sampled node maintains more connections to others within the subgraph.
Panel~(b) shows that CRW outperforms the BFS2 in seed-connected ratio across the board, meaning more nodes in the subgraph are structurally connected to the seed node.
Panel~(c) shows that CRW achieves a substantially larger average component size across most tasks, indicating that its sampled nodes coalesce into fewer, more cohesive clusters rather than scattering across many small fragments.
Finally, panel~(d) illustrates token-budget efficiency, where CRW proves far more effective at packing distinct, structurally relevant nodes into the fixed context window on the vast majority of tasks.
The only exceptions to this efficiency trend are the rel-event tasks, where extreme seed fan-out overwhelms the BFS2 subgraph with a large number of 1-hop neighbors; in contrast, CRW’s hub truncation actively prevents this over-sampling.
Even in these edge cases, CRW still matches BFS2 in core connectivity metrics such as average degree and seed-connected ratio, Figure~\ref{fig:crw-bfs-panel}(a)(b).

Table~\ref{tab:crw-bfs-structure} provides the complete per-task numerical breakdown of these metrics, alongside raw edge density and clustering coefficients.
Notably, raw edge density is a confounded and unreliable comparison metric in this context.
While it mathematically appears to favor BFS2, this is simply an artifact of BFS2 generating smaller subgraphs.

\begin{table*}[t]
\centering
\caption{Per-task subgraph statistics for CRW vs.\ BFS ($K{=}300$, causal edges, all training seeds).
Each cell reports \textbf{CRW\,|\,BFS} means.
Avg.\ Degree, Seed-Conn.\ Ratio, \# Components, and Distinct Nodes are the structurally informative metrics.
Edge Density ($|E|/\binom{n}{2}$) is inflated for BFS by its smaller subgraphs (see text), and
Clustering Coefficient is near-zero on 10/12 tasks due to the realational structure of the underlying graphs.}
\label{tab:crw-bfs-structure}
\resizebox{\textwidth}{!}{%
\begin{tabular}{ll cc cc cc cc cc cc}
\toprule
& & \multicolumn{2}{c}{\textbf{Avg.\ Degree}} & \multicolumn{2}{c}{\textbf{\small Seed-Conn.\ Ratio}} & \multicolumn{2}{c}{\textbf{\# Comp.}} & \multicolumn{2}{c}{\textbf{Distinct Nodes}} & \multicolumn{2}{c}{\textbf{Edge Density}} & \multicolumn{2}{c}{\textbf{Clust.\ Coeff.}} \\
\cmidrule(lr){3-4}\cmidrule(lr){5-6}\cmidrule(lr){7-8}\cmidrule(lr){9-10}\cmidrule(lr){11-12}\cmidrule(lr){13-14}
\textbf{Dataset} & \textbf{Task} & CRW & BFS & CRW & BFS & CRW & BFS & CRW & BFS & CRW & BFS & CRW & BFS \\
\midrule
rel-amazon & item-churn     & 1.99 & 1.80 & 1.000 & 0.988 & 1.1   & 3.3   & 162.2 & 38.3  & 0.032 & 0.181 & 0.000 & 0.000 \\
           & user-churn     & 2.00 & 1.70 & 1.000 & 0.998 & 1.0   & 1.4   & 167.4 & 16.5  & 0.024 & 0.298 & 0.000 & 0.000 \\
\midrule
rel-avito & user-clicks     & 1.98 & 1.50 & 0.646 & 0.595 & 107.2 & 120.2 & 254.2 & 191.6 & 0.009 & 0.049 & 0.000 & 0.000 \\
          & user-visits     & 1.83 & 1.39 & 0.605 & 0.564 & 119.6 & 129.9 & 252.1 & 193.4 & 0.009 & 0.050 & 0.000 & 0.000 \\
\midrule
rel-event & user-ignore     & 1.29 & 1.29 & 0.649 & 0.642 & 106.4 & 101.3 & 177.1 & 273.7 & 0.016 & 0.008 & 0.000 & 0.000 \\
          & user-repeat     & 1.52 & 1.49 & 0.762 & 0.743 & 72.5  & 73.5  & 172.9 & 278.0 & 0.015 & 0.007 & 0.000 & 0.000 \\
\midrule
rel-f1 & driver-dnf        & 2.97 & 2.60 & 0.930 & 0.918 & 22.1  & 24.5  & 269.8 & 136.8 & 0.011 & 0.060 & 0.000 & 0.000 \\
       & driver-top3       & 3.49 & 2.97 & 0.964 & 0.949 & 11.8  & 15.5  & 263.1 & 164.3 & 0.014 & 0.055 & 0.000 & 0.000 \\
\midrule
rel-hm & user-churn         & 2.04 & 1.92 & 0.999 & 0.996 & 1.3   & 1.5   & 235.4 & 32.1  & 0.010 & 0.175 & 0.000 & 0.000 \\
\midrule
rel-stack & user-badge      & 1.29 & 1.26 & 0.652 & 0.648 & 105.5 & 106.3 & 130.9 & 118.5 & 0.235 & 0.289 & 0.108 & 0.119 \\
          & user-engage     & 2.49 & 2.41 & 0.999 & 0.990 & 1.4   & 3.2   & 61.8  & 32.4  & 0.065 & 0.188 & 0.260 & 0.288 \\
\midrule
rel-trial & study-outcome   & 2.33 & 2.17 & 1.000 & 0.989 & 1.0   & 3.2   & 157.1 & 49.4  & 0.017 & 0.107 & 0.003 & 0.000 \\
\midrule
\multicolumn{2}{l}{\textbf{Average (12 tasks)}} & \textbf{2.10} & 1.87 & \textbf{0.850} & 0.835 & \textbf{45.9} & 48.7 & \textbf{190.0} & 126.1 & 0.038 & \textbf{0.122} & 0.031 & \textbf{0.034} \\
\bottomrule
\end{tabular}%
}
\end{table*}

For a subgraph with $n$ unique nodes and $M$ edges, edge density and average degree are defined as
\begin{equation}
  \text{edge\_density} = \frac{M}{\,n(n-1)/2\,},
  \label{eq:edge-density}
\end{equation}
\begin{equation}
  \text{avg\_degree} = \frac{2M}{n}.
  \label{eq:avg-degree}
\end{equation}
Substituting~\eqref{eq:avg-degree} into~\eqref{eq:edge-density} reveals the direct relationship:
\begin{equation}
  \text{edge\_density} = \frac{\text{avg\_degree}}{n - 1}.
  \label{eq:density-degree}
\end{equation}
By effectively dividing the average degree by $n-1$, the quadratic denominator in~\eqref{eq:edge-density} unfairly penalizes the CRW sampler for successfully extracting a much larger number of unique nodes.
Consequently, average degree~\eqref{eq:avg-degree} provides the correct, size-invariant alternative for measuring structural density, ensuring the sampler isn’t penalized simply for recovering a larger number of relevant nodes.

The clustering coefficient is not an appropriate measure of local cohesion for these graphs.
The standard clustering coefficient counts closed triangles among a node’s neighbors, but in most of the heterogeneous relational graphs underlying these benchmarks, closed triangles are structurally impossible under the predefined relational schema.
Accordingly, clustering is identically zero for both samplers on 10 of the 12 tasks.
The appropriate size-invariant measures of local cohesion---average degree and seed-connected ratio, both reported in Table~\ref{tab:crw-bfs-structure}---show clear CRW advantages across all tasks.
On the two \texttt{rel-stack} tasks whose relational schema contains 3-cycles and therefore allows closed triangles, the gap between CRW and BFS2 is minimal ($|d| \le 0.24$), confirming that CRW’s broader exploration does not sacrifice local cohesion.

\paragraph{Summary and downstream impact.}
Ultimately, CRW provides a fundamentally denser and more seed-connected topology than the uniform BFS2 baseline.
This dense connectivity directly improves the expressiveness of the GIN positional encoder and ensures the Local Attention Module attends over a cohesive neighborhood rather than a disjointed set of redundant orphan tokens. Consequently, QUARTET, leveraging CRW, can effectively propagate seed-relevant information while requiring fewer local self-attention layers compared to RelGT (as demonstrated in Figure \ref{fig:rwr-vs-bfs2}).


\subsubsection{Validation and Test Scores for All Baselines}
\label{app:baselines-full}
Table~\ref{tab:baselines-full} reports both validation and test ROC-AUC (mean\,$\pm$\,std) for HGT, RelGT, and \textsc{Quartet} on all twelve classification tasks, complementing the test-only summary of Table~\ref{tab:baselines}.
All four columns are averaged over $n=4$ seeds; the three baseline columns are our own reproductions, and the \textsc{Quartet} column's test rows coincide with Table~\ref{tab:baselines}.
\begin{table}[t]
\centering
\small
\setlength{\tabcolsep}{3pt}
\caption{Validation and test ROC-AUC (mean\,$\pm$\,std) on the twelve \textsc{RelBench} entity-classification tasks for the two reproduced graph-transformer baselines---HGT~\cite{hu2020hgt} and RelGT~\cite{dwivedi2026relgt}---and \textsc{Quartet}. All three columns are averaged over $n=4$ seeds under a shared protocol; HGT and RelGT are our own reproductions. The test rows coincide with Table~\ref{tab:baselines}. Bold marks the best value in each (task, split) row across the three models.}
\label{tab:baselines-full}
\begin{tabular}{lllccc}
\toprule
Dataset & Task & AUC\,$\uparrow$ & HGT & RelGT & \textsc{Quartet} \\
\midrule
rel-amazon & item-churn & Test & 0.7758 {\scriptsize $\pm$ 0.0075} & 0.8249 {\scriptsize $\pm$ 0.0005} & \textbf{0.8262} {\scriptsize $\pm$ 0.0010} \\
           &            & Val  & 0.7776 {\scriptsize $\pm$ 0.0054} & 0.8218 {\scriptsize $\pm$ 0.0004} & \textbf{0.8227} {\scriptsize $\pm$ 0.0007} \\
           & user-churn & Test & 0.6626 {\scriptsize $\pm$ 0.0041} & \textbf{0.7024} {\scriptsize $\pm$ 0.0015} & 0.7001 {\scriptsize $\pm$ 0.0016} \\
           &            & Val  & 0.6664 {\scriptsize $\pm$ 0.0034} & \textbf{0.7023} {\scriptsize $\pm$ 0.0013} & 0.7003 {\scriptsize $\pm$ 0.0008} \\
\midrule
rel-avito & user-clicks & Test & \textbf{0.6501} {\scriptsize $\pm$ 0.0104} & 0.6425 {\scriptsize $\pm$ 0.0159} & 0.6481 {\scriptsize $\pm$ 0.0122} \\
          &             & Val  & 0.5882 {\scriptsize $\pm$ 0.0189} & \textbf{0.6614} {\scriptsize $\pm$ 0.0043} & 0.6087 {\scriptsize $\pm$ 0.0023} \\
          & user-visits & Test & 0.6410 {\scriptsize $\pm$ 0.0068} & \textbf{0.6639} {\scriptsize $\pm$ 0.0020} & 0.6508 {\scriptsize $\pm$ 0.0050} \\
          &             & Val  & 0.6578 {\scriptsize $\pm$ 0.0085} & \textbf{0.6986} {\scriptsize $\pm$ 0.0021} & 0.6892 {\scriptsize $\pm$ 0.0032} \\
\midrule
rel-event & user-ignore & Test & 0.7256 {\scriptsize $\pm$ 0.0080} & 0.7348 {\scriptsize $\pm$ 0.0183} & \textbf{0.7392} {\scriptsize $\pm$ 0.0035} \\
          &             & Val  & 0.7564 {\scriptsize $\pm$ 0.0024} & 0.7691 {\scriptsize $\pm$ 0.0051} & \textbf{0.8176} {\scriptsize $\pm$ 0.0055} \\
          & user-repeat & Test & 0.7539 {\scriptsize $\pm$ 0.0157} & 0.7160 {\scriptsize $\pm$ 0.0316} & \textbf{0.7563} {\scriptsize $\pm$ 0.0175} \\
          &             & Val  & 0.6679 {\scriptsize $\pm$ 0.0013} & 0.7101 {\scriptsize $\pm$ 0.0121} & \textbf{0.7292} {\scriptsize $\pm$ 0.0141} \\
\midrule
rel-f1 & driver-dnf  & Test & 0.6595 {\scriptsize $\pm$ 0.0627} & \textbf{0.7133} {\scriptsize $\pm$ 0.0416} & 0.7044 {\scriptsize $\pm$ 0.0180} \\
       &             & Val  & 0.7756 {\scriptsize $\pm$ 0.0348} & 0.8029 {\scriptsize $\pm$ 0.0086} & \textbf{0.8040} {\scriptsize $\pm$ 0.0041} \\
       & driver-top3 & Test & 0.7384 {\scriptsize $\pm$ 0.0574} & 0.8067 {\scriptsize $\pm$ 0.0402} & \textbf{0.8517} {\scriptsize $\pm$ 0.0156} \\
       &             & Val  & 0.7620 {\scriptsize $\pm$ 0.0633} & \textbf{0.8843} {\scriptsize $\pm$ 0.0118} & 0.8741 {\scriptsize $\pm$ 0.0080} \\
\midrule
rel-hm & user-churn & Test & 0.6746 {\scriptsize $\pm$ 0.0036} & 0.6975 {\scriptsize $\pm$ 0.0016} & \textbf{0.6976} {\scriptsize $\pm$ 0.0011} \\
       &            & Val  & 0.6807 {\scriptsize $\pm$ 0.0041} & \textbf{0.7049} {\scriptsize $\pm$ 0.0012} & 0.7033 {\scriptsize $\pm$ 0.0011} \\
\midrule
rel-stack & user-badge & Test & 0.8688 {\scriptsize $\pm$ 0.0025} & \textbf{0.8764} {\scriptsize $\pm$ 0.0016} & 0.8528 {\scriptsize $\pm$ 0.0094} \\
          &            & Val  & 0.8818 {\scriptsize $\pm$ 0.0018} & \textbf{0.8884} {\scriptsize $\pm$ 0.0010} & 0.8580 {\scriptsize $\pm$ 0.0115} \\
          & user-engagement & Test & 0.8899 {\scriptsize $\pm$ 0.0018} & 0.9047 {\scriptsize $\pm$ 0.0009} & \textbf{0.9063} {\scriptsize $\pm$ 0.0010} \\
          &                 & Val  & 0.8897 {\scriptsize $\pm$ 0.0012} & 0.9010 {\scriptsize $\pm$ 0.0008} & \textbf{0.9029} {\scriptsize $\pm$ 0.0005} \\
\midrule
rel-trial & study-outcome & Test & 0.6345 {\scriptsize $\pm$ 0.0251} & 0.6632 {\scriptsize $\pm$ 0.0177} & \textbf{0.7059} {\scriptsize $\pm$ 0.0044} \\
          &               & Val  & 0.6207 {\scriptsize $\pm$ 0.0303} & \textbf{0.6751} {\scriptsize $\pm$ 0.0073} & 0.6746 {\scriptsize $\pm$ 0.0067} \\
\bottomrule
\end{tabular}
\end{table}

\subsubsection{Full Ablation Results with Seed Variability}
\label{app:ablation-full}
Table~\ref{tab:ablation-full} reports the absolute test ROC-AUC (mean\,$\pm$\,std over $n{=}4$ seeds) for every leave-one-out ablation arm, complementing the relative-change summary of Table~\ref{tab:ablation}.

\clearpage
\pdfpageattr{/Rotate 180}
\begin{landscape}
\vspace*{\fill}
\begin{center}
\footnotesize
\setlength{\tabcolsep}{2.5pt}
\captionof{table}{Full ablation results: test ROC-AUC (mean\,$\pm$\,std, $n{=}4$ seeds) for each leave-one-out ablation of \textsc{Quartet}. Column groups match Table~\ref{tab:ablation}. \colorbox{red!25}{Red cells} denote a statistically significant drop relative to the full \textsc{Quartet} model (i.e., the ablation's upper confidence bound $\mu_{\mathrm{abl}} + \sigma_{\mathrm{abl}}$ falls below the full model's lower bound $\mu_{\mathrm{full}} - \sigma_{\mathrm{full}}$). Bold marks the full \textsc{Quartet} (reference) column.}
\label{tab:ablation-full}
\begin{tabular}{l l c c c @{\hskip 0.8em} c c c @{\hskip 0.8em} c c c c}
\toprule
\multirow{2}{*}{Dataset} & \multirow{2}{*}{Task} & \textbf{\textsc{Quartet}} & No & No & No & No Hub & No & No & No & No & No \\
 & & \textbf{(Full)} & Local & Perceiver & CRW & Trunc. & Restart & Temp. & Collab. & Feat. & Struct. \\
\midrule
rel-amazon
 & item-churn
 & \textbf{.8262 {\scriptsize $\pm$ .0010}}
 & \cellcolor{red!25}.7451 {\scriptsize $\pm$ .0017}
 & \cellcolor{red!25}.8242 {\scriptsize $\pm$ .0007}
 & .8244 {\scriptsize $\pm$ .0008}
 & .8249 {\scriptsize $\pm$ .0008}
 & .8253 {\scriptsize $\pm$ .0011}
 & .8245 {\scriptsize $\pm$ .0009}
 & .8257 {\scriptsize $\pm$ .0008}
 & .8254 {\scriptsize $\pm$ .0009}
 & .8251 {\scriptsize $\pm$ .0013} \\
 & user-churn
 & \textbf{.7001 {\scriptsize $\pm$ .0016}}
 & \cellcolor{red!25}.6629 {\scriptsize $\pm$ .0008}
 & .7011 {\scriptsize $\pm$ .0008}
 & \cellcolor{red!25}.6923 {\scriptsize $\pm$ .0032}
 & .6981 {\scriptsize $\pm$ .0016}
 & \cellcolor{red!25}.6954 {\scriptsize $\pm$ .0006}
 & .6996 {\scriptsize $\pm$ .0017}
 & .7007 {\scriptsize $\pm$ .0011}
 & .6976 {\scriptsize $\pm$ .0025}
 & \cellcolor{red!25}.6962 {\scriptsize $\pm$ .0022} \\
\midrule
rel-avito
 & user-clicks
 & \textbf{.6481 {\scriptsize $\pm$ .0122}}
 & \cellcolor{red!25}.5952 {\scriptsize $\pm$ .0243}
 & .6348 {\scriptsize $\pm$ .0299}
 & .6534 {\scriptsize $\pm$ .0034}
 & .6522 {\scriptsize $\pm$ .0068}
 & .5826 {\scriptsize $\pm$ .0847}
 & .6410 {\scriptsize $\pm$ .0081}
 & \cellcolor{red!25}.6224 {\scriptsize $\pm$ .0128}
 & .6111 {\scriptsize $\pm$ .0460}
 & .6367 {\scriptsize $\pm$ .0231} \\
 & user-visits
 & \textbf{.6508 {\scriptsize $\pm$ .0050}}
 & \cellcolor{red!25}.6409 {\scriptsize $\pm$ .0046}
 & .6506 {\scriptsize $\pm$ .0032}
 & .6428 {\scriptsize $\pm$ .0136}
 & .6508 {\scriptsize $\pm$ .0034}
 & .6450 {\scriptsize $\pm$ .0097}
 & .6435 {\scriptsize $\pm$ .0062}
 & \cellcolor{red!25}.6293 {\scriptsize $\pm$ .0157}
 & .6458 {\scriptsize $\pm$ .0160}
 & .6333 {\scriptsize $\pm$ .0202} \\
\midrule
rel-event
 & user-ignore
 & \textbf{.7392 {\scriptsize $\pm$ .0035}}
 & .7740 {\scriptsize $\pm$ .0107}
 & .7334 {\scriptsize $\pm$ .0050}
 & .7336 {\scriptsize $\pm$ .0279}
 & .7392 {\scriptsize $\pm$ .0139}
 & .7380 {\scriptsize $\pm$ .0048}
 & .7272 {\scriptsize $\pm$ .0100}
 & \cellcolor{red!25}.7270 {\scriptsize $\pm$ .0075}
 & .7295 {\scriptsize $\pm$ .0151}
 & .7410 {\scriptsize $\pm$ .0098} \\
 & user-repeat
 & \textbf{.7563 {\scriptsize $\pm$ .0175}}
 & \cellcolor{red!25}.5642 {\scriptsize $\pm$ .0664}
 & .7425 {\scriptsize $\pm$ .0168}
 & .7099 {\scriptsize $\pm$ .0615}
 & .6780 {\scriptsize $\pm$ .1072}
 & .7388 {\scriptsize $\pm$ .0104}
 & .7283 {\scriptsize $\pm$ .0228}
 & .7450 {\scriptsize $\pm$ .0235}
 & .7359 {\scriptsize $\pm$ .0327}
 & .7442 {\scriptsize $\pm$ .0244} \\
\midrule
rel-f1
 & driver-dnf
 & \textbf{.7044 {\scriptsize $\pm$ .0180}}
 & \cellcolor{red!25}.6743 {\scriptsize $\pm$ .0105}
 & .6641 {\scriptsize $\pm$ .0337}
 & .7015 {\scriptsize $\pm$ .0169}
 & .7103 {\scriptsize $\pm$ .0119}
 & .7059 {\scriptsize $\pm$ .0155}
 & .6714 {\scriptsize $\pm$ .0292}
 & .6847 {\scriptsize $\pm$ .0183}
 & .6789 {\scriptsize $\pm$ .0236}
 & .6803 {\scriptsize $\pm$ .0086} \\
 & driver-top3
 & \textbf{.8517 {\scriptsize $\pm$ .0156}}
 & \cellcolor{red!25}.7074 {\scriptsize $\pm$ .0161}
 & \cellcolor{red!25}.7925 {\scriptsize $\pm$ .0175}
 & \cellcolor{red!25}.8223 {\scriptsize $\pm$ .0039}
 & \cellcolor{red!25}.7758 {\scriptsize $\pm$ .0241}
 & \cellcolor{red!25}.8176 {\scriptsize $\pm$ .0177}
 & \cellcolor{red!25}.8130 {\scriptsize $\pm$ .0200}
 & .8201 {\scriptsize $\pm$ .0479}
 & .8403 {\scriptsize $\pm$ .0182}
 & .8247 {\scriptsize $\pm$ .0288} \\
\midrule
rel-hm
 & user-churn
 & \textbf{.6976 {\scriptsize $\pm$ .0011}}
 & \cellcolor{red!25}.6748 {\scriptsize $\pm$ .0053}
 & .6970 {\scriptsize $\pm$ .0028}
 & .6961 {\scriptsize $\pm$ .0021}
 & .6988 {\scriptsize $\pm$ .0014}
 & .6964 {\scriptsize $\pm$ .0012}
 & \cellcolor{red!25}.6949 {\scriptsize $\pm$ .0013}
 & .6967 {\scriptsize $\pm$ .0018}
 & .6957 {\scriptsize $\pm$ .0012}
 & .6953 {\scriptsize $\pm$ .0018} \\
\midrule
rel-stack
 & user-badge
 & \textbf{.8528 {\scriptsize $\pm$ .0094}}
 & .8182 {\scriptsize $\pm$ .0707}
 & \cellcolor{red!25}.7859 {\scriptsize $\pm$ .0147}
 & .8339 {\scriptsize $\pm$ .0170}
 & .8262 {\scriptsize $\pm$ .0194}
 & \cellcolor{red!25}.8122 {\scriptsize $\pm$ .0185}
 & \cellcolor{red!25}.8163 {\scriptsize $\pm$ .0182}
 & \cellcolor{red!25}.8157 {\scriptsize $\pm$ .0258}
 & .8275 {\scriptsize $\pm$ .0292}
 & .8319 {\scriptsize $\pm$ .0137} \\
 & user-engagement
 & \textbf{.9063 {\scriptsize $\pm$ .0010}}
 & \cellcolor{red!25}.8644 {\scriptsize $\pm$ .0024}
 & .9064 {\scriptsize $\pm$ .0007}
 & .9056 {\scriptsize $\pm$ .0011}
 & .9059 {\scriptsize $\pm$ .0010}
 & .9051 {\scriptsize $\pm$ .0012}
 & .9054 {\scriptsize $\pm$ .0004}
 & .9043 {\scriptsize $\pm$ .0023}
 & .9038 {\scriptsize $\pm$ .0018}
 & .9047 {\scriptsize $\pm$ .0025} \\
\midrule
rel-trial
 & study-outcome
 & \textbf{.7059 {\scriptsize $\pm$ .0044}}
 & .6925 {\scriptsize $\pm$ .0094}
 & .7014 {\scriptsize $\pm$ .0100}
 & \cellcolor{red!25}.6804 {\scriptsize $\pm$ .0191}
 & .7038 {\scriptsize $\pm$ .0057}
 & \cellcolor{red!25}.6930 {\scriptsize $\pm$ .0076}
 & \cellcolor{red!25}.6931 {\scriptsize $\pm$ .0041}
 & .7045 {\scriptsize $\pm$ .0204}
 & .6974 {\scriptsize $\pm$ .0085}
 & \cellcolor{red!25}.6889 {\scriptsize $\pm$ .0034} \\
\bottomrule
\end{tabular}
\end{center}
\vspace*{\fill}
\end{landscape}
\clearpage
\pdfpageattr{}

\subsubsection{RGP Baseline Reproduction}
\label{app:rgp-reproduction}
The Relational Graph Perceiver (RGP)~\cite{lachi2025rgp} is the most architecturally related prior work to \textsc{Quartet}, as both compress structural and temporal context through Perceiver-style cross-attention.
Because RGP does not have publicly available code for running benchmark comparisons, we performed a ground-up reproduction following direct guidance from the original authors.
Our reproduction assembles RGP's full forward path from three verified sources:
\begin{itemize}[nosep]
  \item \textbf{Tokenization:} RelGT's structural tokenizer (feature encoding via \texttt{NeighborTfsEncoder}) combined with the RGP paper's own positional-encoding formula (type $+$ centrality $+$ hop $+$ time), which the author confirmed differs from RelGT's.
  \item \textbf{Perceiver core:} GraphFM~\cite{lachi2026graphfmgeneralistgraphtransformer}'s memory-efficient cross-attention and feed-forward blocks (from the same research group as the RGP authors) reused as the dual-branch encoder backbone, as confirmed by the author.
  \item \textbf{Temporal sampler:} Implemented from scratch following the paper's Algorithm~1 pseudocode. We use the selection rule (per-edge-type top-$k{=}10$, strictly causal, no time-window mode) as confirmed by the author.
\end{itemize}
Because the original paper omitted per-task hyperparameters, we performed a grid search over Perceiver layers $L \in \{2,4,6\}$ and latent tokens $n \in \{8,16,32\}$ for each task.

Table~\ref{tab:rgp-reproduction} presents the completed reproduction results across all twelve \textsc{RelBench} classification tasks. Overall, \textsc{Quartet} maintains highly competitive performance in this expanded baseline set, achieving the highest mean ROC-AUC on six tasks, compared to four for RelGT and two for RGP.
We restrict these results to the appendix rather than incorporating them into the main text (Table~\ref{tab:baselines}) due to remaining methodological ambiguities that prevent a guaranteed apples-to-apples comparison. Specifically, without access to the original codebase, we cannot definitively rule out the use of undocumented mechanisms such as temporal decay, nor can we guarantee that the Perceiver architecture ported from GraphFM is strictly identical to the proprietary RGP model. Given these constraints, the results in Table~\ref{tab:rgp-reproduction} are provided as a transparent, best-effort reproduction.

\begin{table}[t]
\centering
\small
\setlength{\tabcolsep}{3pt}
\caption{RGP reproduction: test ROC-AUC (mean\,$\pm$\,std) on the twelve \textsc{RelBench} classification tasks. RGP is reproduced following direct guidance from the original authors (\S\ref{app:rgp-reproduction}); HGT and RelGT results are from Table~\ref{tab:baselines}. All results are averaged over $n=4$ seeds. \textbf{Bold} marks the best mean in each row.}
\label{tab:rgp-reproduction}
\begin{tabular}{llcccc}
\toprule
Dataset & Task & RGP & HGT & RelGT & \textsc{Quartet} \\
\midrule
rel-amazon & item-churn      & 0.7641 {\scriptsize $\pm$ 0.0093} & 0.7758 {\scriptsize $\pm$ 0.0075} & 0.8249 {\scriptsize $\pm$ 0.0005} & \textbf{0.8262} {\scriptsize $\pm$ 0.0004} \\
           & user-churn      & 0.6283 {\scriptsize $\pm$ 0.0041} & 0.6626 {\scriptsize $\pm$ 0.0041} & \textbf{0.7024} {\scriptsize $\pm$ 0.0015} & 0.7008 {\scriptsize $\pm$ 0.0010} \\
\midrule
rel-avito  & user-clicks     & \textbf{0.6515} {\scriptsize $\pm$ 0.0033} & 0.6501 {\scriptsize $\pm$ 0.0104} & 0.6425 {\scriptsize $\pm$ 0.0159} & 0.6482 {\scriptsize $\pm$ 0.0128} \\
           & user-visits     & 0.6534 {\scriptsize $\pm$ 0.0009} & 0.6410 {\scriptsize $\pm$ 0.0068} & \textbf{0.6639} {\scriptsize $\pm$ 0.0020} & 0.6508 {\scriptsize $\pm$ 0.0076} \\
\midrule
rel-event  & user-ignore     & \textbf{0.7765} {\scriptsize $\pm$ 0.0312} & 0.7278 {\scriptsize $\pm$ 0.0061} & 0.7348 {\scriptsize $\pm$ 0.0183} & 0.7397 {\scriptsize $\pm$ 0.0037} \\
           & user-repeat     & 0.6849 {\scriptsize $\pm$ 0.0520} & 0.7581 {\scriptsize $\pm$ 0.0120} & 0.7160 {\scriptsize $\pm$ 0.0316} & \textbf{0.7603} {\scriptsize $\pm$ 0.0351} \\
\midrule
rel-f1     & driver-dnf      & 0.7071 {\scriptsize $\pm$ 0.0184} & 0.6595 {\scriptsize $\pm$ 0.0627} & \textbf{0.7133} {\scriptsize $\pm$ 0.0416} & 0.7045 {\scriptsize $\pm$ 0.0131} \\
           & driver-top3     & 0.7815 {\scriptsize $\pm$ 0.0415} & 0.7384 {\scriptsize $\pm$ 0.0574} & 0.8067 {\scriptsize $\pm$ 0.0402} & \textbf{0.8522} {\scriptsize $\pm$ 0.0242} \\
\midrule
rel-hm     & user-churn      & 0.6777 {\scriptsize $\pm$ 0.0004} & 0.6746 {\scriptsize $\pm$ 0.0036} & 0.6975 {\scriptsize $\pm$ 0.0016} & \textbf{0.6976} {\scriptsize $\pm$ 0.0017} \\
\midrule
rel-stack  & user-badge      & 0.8284 {\scriptsize $\pm$ 0.0064} & 0.8703 {\scriptsize $\pm$ 0.0025} & \textbf{0.8764} {\scriptsize $\pm$ 0.0016} & 0.8561 {\scriptsize $\pm$ 0.0131} \\
           & user-engagement & 0.8535 {\scriptsize $\pm$ 0.0153} & 0.8899 {\scriptsize $\pm$ 0.0018} & 0.9047 {\scriptsize $\pm$ 0.0009} & \textbf{0.9064} {\scriptsize $\pm$ 0.0003} \\
\midrule
rel-trial  & study-outcome   & 0.6814 {\scriptsize $\pm$ 0.0092} & 0.6345 {\scriptsize $\pm$ 0.0251} & 0.6632 {\scriptsize $\pm$ 0.0177} & \textbf{0.7058} {\scriptsize $\pm$ 0.0054} \\
\bottomrule
\end{tabular}
\end{table}

\end{document}